\documentclass[]{jdsart}
\usepackage{amsthm,amsmath,amssymb}
\newcommand\mbb[1]{\mathbb{#1}}

\def\reals{\mathbb{R}} 
\renewcommand{\exp}[1]{\operatorname{exp}\left(#1\right)} 
\def\indic#1{\mbb{I}\left({#1}\right)} 

\def\Dir{\textnormal{Dir}}

\def\*#1{\mathbf{#1}}

\usepackage{listings}
\usepackage{array}
\newcolumntype{R}[1]{>{\raggedleft\let\newline\\\arraybackslash\hspace{0pt}}m{#1}}

\volume{0}
\issue{0}
\pubyear{2022}
\articletype{research-article}
\doi{0000}

\begin{document}
\begin{frontmatter}
\pretitle{Research Article}
\title{Data Science Principles for Interpretable and Explainable AI}
\runtitle{Interpretable and Explainable AI}
\author[1]{\inits{K} \fnms{Kris} \snm{Sankaran}\thanksref{c1}\ead{ksankaran@wisc.edu}}
\thankstext[type=corresp,id=c1]{Corresponding author.}
\address[1]{Department of Statistics \\ \institution{University of Wisconsin - Madison}, \cny{United States}}

\begin{abstract}
Society's capacity for algorithmic problem-solving has never been greater.
Artificial Intelligence is now applied across more domains than ever, a
consequence of powerful abstractions, abundant data, and accessible
software. As capabilities have expanded, so have risks, with models often
deployed without fully understanding their potential impacts. Interpretable
and interactive machine learning aims to make complex models more transparent
and controllable, enhancing user agency. This review synthesizes key principles
from the growing literature in this field.
We first introduce precise vocabulary for discussing interpretability, like the
distinction between glass box and explainable models. We then explore
connections to classical statistical and design principles, like parsimony and
the gulfs of interaction. Basic explainability techniques -- including learned
embeddings, integrated gradients, and concept bottlenecks -- are illustrated
with a simple case study. We also review criteria for objectively evaluating
interpretability approaches. Throughout, we underscore the importance of
considering audience goals when designing interactive data-driven systems.
Finally, we outline open challenges and discuss the potential role of data
science in addressing them. Code to reproduce all examples can be found at
\href{https://go.wisc.edu/3k1ewe}{https://go.wisc.edu/3k1ewe}.
\end{abstract}

\end{frontmatter}

\section{Introduction}
\label{sec:introduction}
The success of Artificial Intelligence (AI) stems in part from its ability to
abstract away details of its context. When a model is applied to predict
responses from a collection of features, it matters little whether the response
is the shape of a galaxy or the next word in a sentence. Such abstraction has
enabled the design of versatile methods with remarkably diverse applications.
Nonetheless, models are never used in a vacuum. Models affect human
well-being, and it is important to systematically study the human-model
interface. How can we ensure that a model behaves acceptably in unforeseen
circumstances? How can we prevent harm and unfair treatment? How can we align model behavior with our
expectations? The field of interpretable machine learning and explainable AI
(XAI) has emerged to address these questions, allowing
model-based abstractions to be used wisely within messy, real-world
contexts. This literature has engaged researchers from various backgrounds, from
theorists seeking to characterize the fundamental limits of explainability
techniques \citep{pmlr-v119-williamson20a, Bilodeau2024} to philosophers and
psychologists critiquing popular discourses surrounding them
\citep{Krishnan2019, Nussberger2022}.

What can go wrong without interpretability? Consider these examples:
\begin{itemize}
\item \cite{10.1145/2783258.2788613} aimed to help a hospital triage incoming
pneumonia patients, ensuring that high-risk patients are admitted while allowing
lower-risk patients to receive outpatient care. Using a dataset of past
pneumonia patients, their model achieved an area under the receiver operating
characteristic curve (AUC) of 0.857 in predicting patient mortality based on lab
and medical profiles.  Surprisingly, an important feature analysis revealed that
asthmatic patients had lower predicted probability of death.  This arose because
the original training data were observational. Historically, asthmatic patients
were treated more aggressively than those without asthma, and therefore enjoyed
lower mortality. Directly deploying such a model would put asthma patients at
risk. Fortunately, their model was editable, so the asthma association could be
removed before deployment. \item \cite{Gu2019} designed stickers that, when
attached to stop signs, caused object detector models used by self-driving
cars to misclassify them as speed limits. This real-world adversarial attack
exploits the sensitivity of models to small (and, to humans, seemingly
arbitrary) input perturbations. More interpretable models could yield systems
with more predictable behavior. \item In Google’s first public demo of the
generative AI chatbot Bard, it was asked, ``What new discoveries from the James
Webb Space Telescope (JWST) can I tell my 9 year old about?'' The chatbot
responded with three seemingly legitimate facts. Only after the demo was shared
online did astronomers point out that the third fact, ``JWST took the very first
pictures of a planet outside of our own solar system,'' was in fact untrue — the
first exoplanet had been imaged in 2004 \citep{kundaliya2023}. AI chatbots can
``hallucinate'' falsehoods, even when their output seems authoritative. Research
on XAI could shed light on how these systems generate responses, allowing them
to be treated less like oracles and more like search engines. \end{itemize}

These examples beg the question: What makes a model interpretable?
Directly answering this questions is complex, so let's consider a simpler one:
What makes a data visualization effective? This question has been studied for
decades \citep{Cleveland1993-pd, Tufte2001-xs, 10.1145/1924421.1924439, 6327248}, and every data scientist has firsthand experience with it. A
good visualization streamlines a taxing cognitive operation into a perceptual
one. It helps when the visual encoding -- the mapping from sample properties to
graphical elements
-- uses representations that are already familiar or easily learnable. Further,
the graphical elements must be legible and well-annotated. We also tend to learn
more from information-dense visualizations \citep{Tufte2001-xs, 9555620}, since they
prevent oversummarization and can highlight details for follow-up study.
Similarly, an interpretable model can be broken down into relevant components,
each of which can be assigned meaning. Instead of information density (showing
more of the data), interpretability relies on faithfulness (showing more of the
model).

The parallels run deeper. As in a good visualization, the data provenance of a
trustworthy model can be traced back to the original measurement mechanism —
beautiful design and high-accuracy have little value otherwise. Moreover, like
visualization, interpretability must be tailored to an audience and the 
problems they need solved \citep{Lipton2018}. There are different levels of data
literacy, and visual representations may be familiar to some audiences but not
others. Similarly, AI models are employed across a range of problem domains,
necessitating validation in realistic settings (see Section
\ref{sec:evaluation}). Finally, effective visualizations push readers beyond
passive consumption — they inspire deeper exploration of complexity. Likewise,
interpretability can support the transition from automated to augmented
decision-making \citep{Heer2019}, enhancing rather than substituting human
reason.

Model interpretability can be approached with the same nuance that is
already routine in data visualization. The data visualization literature has
identified systematic design approaches and formal evaluation criteria,
cautioning against blanket statements about entire approaches. Indeed, even
information density has been thoughtfully critiqued \citep{6634103}.
Similarly, discussions of interpretability can go beyond dichotomies about
models being either glass or black boxes. To help us navigate the gray areas of
this growing field, Section \ref{sec:vocabulary} introduces relevant vocabulary.
Section \ref{sec:methods} outlines representative techniques and applies them to
a simple simulation example. This section, and the code that accompanies it
(\href{https://go.wisc.edu/3k1ewe}{https://go.wisc.edu/3k1ewe}), can serve as a
tutorial on the practical implementation of interpretability techniques.
Finally, Sections \ref{sec:evaluation} – \ref{sec:discussion} examine the
conceptual and technical questions required for effective evaluation and
progress as the AI landscape evolves.

\subsection{Vocabulary}
\label{sec:vocabulary}

The terms used in interpretability research can be confusing, because though
terms like ``explainable'' are familiar in a colloquial sense, they have a precise
technical meaning in the literature. Indeed, one difficulty is that many people
agree that model interpretability is important — they just have different
ideas about what it looks like. Therefore, it helps to have vocabulary that
distinguishes key properties while maintaining an appropriate level of
abstraction.

An important distinction is between an \emph{intrinsically interpretable model} and an \emph{explainability
technique} \citep{Murdoch2019, Rudin2019}. An intrinsically interpretable model is one that, by
virtue of its design, is easy to accurately describe and alter. For this reason,
they are often called glass boxes. A canonical example is a sparse linear model.
For any new example, a prediction can be constructed in a single pass over the
features with nonzero coefficients. In contrast, explainability techniques are
designed to improve our mental models of arbitrary black box models. A common
example is the partial dependence profile \citep{Friedman2001}, defined as $\mathrm{PDP}\left(x_{d}\right) = \frac{1}{N}\sum_{i = 1}^{N} \hat{f}\left(x_{d}, \*x_{i\left(-d\right)}\right)$, which describes the influence of the $d^{th}$ feature when holding all others fixed at values observed in the original data $\*x_{1}, \dots, \*x_{N}$. Though originally
designed to summarize gradient boosting fits $\hat{f}$, the fact that it only
requires $\left(\*x, \hat{f}\left(\*x\right)\right)$ pairs means it applies
to any supervised learner. To summarize, interpretable models are like glass boxes whose inner
workings are transparent, while explainability techniques are systematic ways
of analyzing the outputs emerging from black boxes.

We can relate these approaches to data science and design principles; see 
Table \ref{tab:principles}. First, consider intrinsically interpretable models. When we call a
linear model interpretable, we are invoking \emph{parsimony} and \emph{simulatability}.
Parsimony means that predictions can be traced back to a few model components,
each of which comes with a simple story attached. For example, sparse linear
models have few nonzero coefficients, and the relationship between each
coefficient and the output can be concisely described. A similar principle
applies to generalized additive modeling, which have few allowable interactions
\citep{10.1145/2783258.2788613}, or in latent variable models, which have few
underlying factors \citep{Sankaran2023}. Simulatability formalizes the idea that
predictions can be manually reconstructed from model descriptions. For example,
in decision trees and falling rules lists, this can be done by answering a
sequence of yes-no questions. Finally, since interpretation requires human
interaction, it can be affected by the gulfs of interaction
\citep{Hutchins1985}. Ideally, an interpretability method should allow users to
quickly query properties of the underlying model. Further, the method's outputs
should be immediately relevant to the query, requiring no further cognitive
processing. To the extent that a method has reached these two ideals, it has reduced
the gulfs of execution and evaluation, respectively.

Unfortunately, even when restricting focus to glass box models, these
definitions are still somewhat ambiguous, leading to issues with evaluation. For example, simulatability alone does not quantify the user
effort required to form predictions; it also ignores their accuracy. Linear
models become unwieldy when they have many nonzero coefficients \citep{Poursabzi-Sangdeh2021-ws}, and deep
decision trees take time to parse. Similarly, latent variables are only
interpretable if they can be related to existing knowledge.

Among explainability techniques, a key question is scale, and this leads to
the distinction between \emph{global} and \emph{local explanations} \citep{Lundberg2020-cu}. Global explanations
summarize the overall structure of a model. Partial dependence profiles are an
example of a global explanation, because they describe $\hat{f}$ across the
range of possible inputs. In contrast, local explanations support reasoning
about an individual sample's prediction. For example, when a sentiment analysis model
classifies a particular hotel review as negative, we might be curious about
which phrases were most important in that classification. Global explanations
shed light on the high-level properties of a model while local explanations 
dissect specific instances of a model's prediction. Local explanations are built
on the observation that perturbations, either of the model or the data, provide
valuable context for understanding model-based decisions. Sensitivity, a
principle with a long history in statistics \citep{Tukey:1959:SSC, Huber1964,
Holmes2017}, is a fruitful path to explainability. Nonetheless, we
caution that differences between local and global can blur in practice \citep{Dandl2023-xl, Herbinger2022-jb}. For
example, concept bottleneck models (Section
\ref{sec:concept_bottleneck_models}), globally restrict the space of possible
prediction functions, forcing them to pass through an interpretable concept
bottleneck. However, editing the bottleneck for specific examples allows for 
local interventions. As detailed in Section \ref{sec:concept_bottleneck_models}, if sensitive attributes are used as concepts, then their coefficients can be set to zero, ensuring that they don't influence the final prediction, even indirectly through related features.
In light of these ambiguities, it is helpful to think of the
vocabulary in this section as highlighting the main axes of variation in the space of
techniques, a map to guide application and evaluation of
interpretability methods.

\begin{table}[]
  \centering
  \begin{tabular}{|l|p{12cm}|}
  \hline
  \textbf{Principle} & \textbf{Discussion} \\ \hline
  Parsimony & To facilitate interpretation, the total number of relevant model
  components should be relatively small. For example, in $\ell_{1}$-regularized
  methods, the number of coefficients is kept small, and decision trees with
  fewer splits are more interpretable than those with many. \\ \hline
  Simulatability & Given a sample and a model description, how easy would it be
  for a user to manually derive the associated prediction?  Regression models
  with few coefficients and approaches like rule lists tend to be more
  simulatable than those that involve more intensive or multistep pooling of
  evidence. \\ \hline
  Sensitivity & How robust are predictions to small changes in either the data
  or the model? Ideas from robust statistics can inform how we understand
  variable importance and stability in more complex AI models. \\ \hline
  Navigating Scales & Local interpretations concern the role of individual
  samples in generating predictions, while global interpretations contribute to
  understanding a model overall. Interpretation at both scales can guide the
  appropriate real-world model use. \\ \hline
  Interaction Gulfs & The ``Gulf of Execution'' is the time it takes for an
  interactive program to respond to a user's input, while the ``Gulf of
  Evaluation'' is the time it takes for the user to understand the output and
  specify a new change. Both can influence the practicality of an
  interpretability method. \\ \hline
  \end{tabular}
  \caption{The core data science principles underlying interpretability. They
  can guide the development of more trustworthy and transparent AI models.}
  \label{tab:principles}
\end{table}

\section{Methods}
\label{sec:methods}

How is interpretability put into practice? We next review approaches 
for direct interpretability and XAI. Rather than attempting to review
all proposals from this rapidly growing literature, we prioritize the simple but
timeless ideas that lie behind larger classes of methods. To ground the
discussion, we include a simulation example where the generative mechanism is
simple to communicate, but complex enough to warrant analysis with AI.

\subsection{Simulation design}
\label{subsec:simulation_design}

The simulation is inspired by the types of longitudinal studies often
encountered in microbiome data science \citep{block_bootstrap, Silverman2018, Kodikara2022}. Consider following 500 participants for
two months, gathering microbiome samples each day. At the study's conclusion,
each participant has their health checked, and they are assigned to either
healthy or disease groups. For example, the participants may have contracted HIV
or developed Type I Diabetes, as in \citep{GOSMANN201729} and
\citep{Kostic2015}, respectively. The question of interest is whether there are
systematic differences in the microbiome trajectories between groups. These
might have diagnostic potential and could shed light on the underlying
mechanisms of disease development. For example, does the disease group have
lower levels of a protective species? Do they exhibit unstable dynamics for any
important microbial subcommunities? For either fundamental biology or diagnostic
potential, any models applied to the data must be either interpretable or explainable
-- classification alone has little value unless it can be incorporated into a
medically coherent narrative. 

\begin{figure}
  \includegraphics[width=0.9\textwidth]{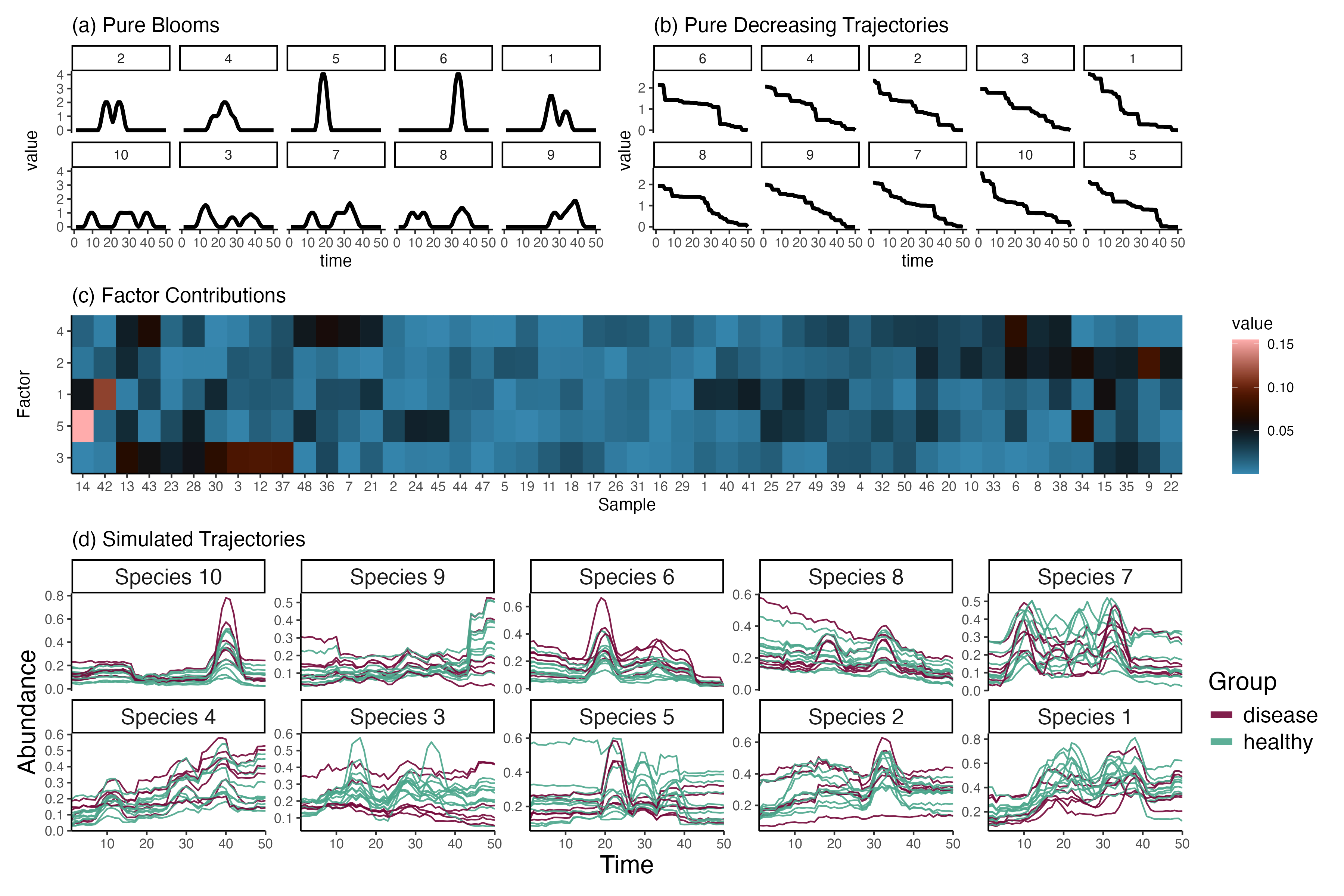}
  \caption{Example simulated trajectories. (a - b). Trajectories for two latent
  types. Each sample is a mixture of these types. Even for a fixed type, samples
  may have different variants, e.g., blooms at different times. (c) Observed
  trajectories are mixtures of these pure underlying trajectory types. The
  contributions of five latent trajectory patterns for each sample are shown
  in this heatmap. (d) Healthy and disease samples generated through this
  process for a subset of samples and taxa.}
  \label{fig:example_trajectories}
\end{figure}

In our hypothetical scenario, we simulate $N \in \{500, 1000, 5000, 10000\}$ participants across $T = 50$
time points, with each sample containing abundances of $D = 144$ microbial
species. Figure \ref{fig:example_trajectories} illustrates trajectories for a
selection of taxa and participants. The generative mechanism is as follows: we
create $K = 25$ community trajectories $\*C_k \in \mathbb{R}^{T \times D}$, for $k
= 1, \dots, K$ by sampling:
\begin{align}
 \*C_{k,d} = \begin{cases}
  \text{Unif}\left[0, 0.01\right]^{T} & \text{ with probability } 0.7 \\
  \*f^{\text{increase}}_{kd} & \text{ with probability } 0.1 \\
  \*f^{\text{decrease}}_{kd} & \text{ with probability } 0.1 \\
  \*f^{\text{bloom}}_{kd} & \text{ with probability } 0.1.
 \end{cases}
 \label{eq:random_trajectory}
\end{align}
The species-wise random trajectories $\*f_{kd}$ are defined as follows.  For
random increases, we compute the cumulative sum of random nonnegative weights.
Decreases, as in Fig \ref{fig:example_trajectories}b, follow the same process
but with a sign reversal. Specifically, the $T$ coordinates of
$\*f_{kd}^{\text{increase}}$ are defined using,
\begin{align}
  \tilde{f}_{kdt}^{\text{increase}} &= \sum_{t' = 1}^{t} u_{t'} \\
  \left(u_1, \dots, u_T\right) &\sim \text{Dirichlet}\left(\lambda_{u}\*1_{T}\right),
\end{align}
and then renormalizing $\tilde{\*f}_{kd}$ to sum to $T$. We set $\lambda_{u} =
0.3$ to allow for occasional large jumps. Random blooms
$\*f_{kd}^{\text{bloom}}$ like those in Fig \ref{fig:example_trajectories}a are
formed using,
\begin{align}
  \tilde{f}_{kdt}^{\text{bloom}} &= \sum_{t^{\ast}} K_{r, L} \left(t - t^{\ast}\right) \\
  t_{1}^{\ast}, \dots, t_{n_{\text{bloom}}}^{\ast} \vert n_{\text{bloom}} &\sim \text{Unif}\left[L, T - L\right] \\
  n_{\text{bloom}} &\sim \text{Poi}\left(\lambda_{\text{bloom}}\right),
\end{align}
where $K_{r}$ is a Tukey window function \citep{gsignal} with bandwidth $r =
0.9$ and window size $L = 9$. The trajectory is then renormalized as with
$\*f_{kd}^{\text{increase}}$. Given dictionaries $\*C_{k}$ of community
trajectories, samples $\*x_{i} \in \reals^{T \times D}$ for subject $i$ are
sampled,
\begin{align}
 \*x_{i} &= \sum_{k = 1}^{K} \theta_{ik} \*C_{k} \\
 \left(\theta _{i1}, \dots, \theta_{i K}\right) &\sim \Dir\left(\lambda_{\theta}\*1_{K}\right).
\end{align}
To generate class assignments $y_{i}$, we cluster the community weights
$\theta_{ik}$ into 24 clusters. These clusters
are then randomly divided into 12 healthy and 12 disease groups. Note that each cluster contains a mixture of the $K = 25$ community trajectories. This approach
therefore links subjects with similar underlying community trajectories into similar
health outcomes in a way that is highly nonlinear.

In practice, the steps of the generative mechanism would not be known. Visually
examining example trajectories in Figure \ref{fig:example_trajectories}d does
not suggest clear choices about which microbiome features are relevant for
disease classification. Each taxon exhibits a complex, nonlinear trend over
time, and while some participants show ``blooms'' at similar times, it's unclear
whether these are linked to disease. Further, the sheer number of taxa makes
drawing conclusions through manual inspection impractical. Computational
approaches are necessary to distinguish disease from healthy trajectories and
identify taxonomic or temporal features of interest, which could then be
validated in follow-up studies involving direct interventions on features in a
controlled setting.

\subsection{Intrinsically interpretable models}
\label{subsec:directly_interpretable_models}

We can often learn salient, non-obvious characteristics of the data by applying
a intrinsically interpretable model. Some have argued that with sufficient ingenuity,
such models can rival the accuracy and efficiency of any black box
\citep{Rudin2019}. Even otherwise, the results of intrinsically interpretable
analyses can guide the application of more sophisticated models later on. With
this in mind, we briefly review sparse logistic regression \citep{Friedman2010}
and decision trees \citep{Loh2014}, apply them to the simulation, and interpret
the results.
\subsubsection{Sparse logistic regression}
Sparse logistic regression solves the optimization problem,
\begin{align*}
\hat{\*\beta} &= \arg\min_{\*\beta \in \mathbb{R}^{D}} \sum_{i = 1}^{N} \ell\left(y_{i} \vert \*x_{i}, \*\beta\right) + \lambda \|\*\beta\|_{1},
\end{align*}
where $\ell\left(y_{i} \vert \*x_{i}, \*\beta\right) = \log
\left(\left(1 + \exp{\*x_{i}^{T}\*\beta y_{i}}\right)^{-1}\exp{\*x_{i}^{T}\*\beta y_{i}}\right)$.
This optimization identifies a sparse coefficient vector $\hat{\*\beta}$
that maximizes the log-likelihood of pairs $\left(\*x_{i}, y_{i}\right)$. The
parameter $\lambda$ controls sparsity: higher values yield simpler but less
expressive models. The combination of sparsity and linearity ensures
interpretability. Sparsity allows many features within $\*x_{i}$ to be ignored
during prediction, while linearity ensures that when features do contribute,
they enter the model transparently. Specifically, a one-unit increase in
the $d^{th}$-coordinate $x_{d}$ changes the relative risk for class $y = 1$ by $\exp{\beta_{d}}$.

\subsubsection{Decision trees}
In contrast, decision trees do not solve a single optimization problem.
Instead, they are defined through a recursive algorithm. Initially, we scan all
$D$ dimensions of $\*x_{i}$ and determine thresholds $t_d$ such that
$\indic{x_{id} > t_{d}}$ maximizes classification accuracy.  From these
candidates, we select the optimal $d^{\ast}$, defining the first partition along
the axis $d^{\ast}$.  The two elements of the partition correspond to two leaves of
a tree split according to $x_{i d^{\ast}}$. Next, for a tree with $L$
leaves, we iteratively split each element along axis $d$ at a threshold
$t_{d}^{l}$, for all pairs of leaves $l$ and dimensions $d$, selecting the split
that maximizes accuracy. This process stops once model complexity outweighs
performance on a holdout set.

Interpretability here stems from parsimony and simulatability. The final tree
has only as many partition elements as leaves, and with sufficient cost on model
complexity, this can be easily visualized as a single tree. Further, the
axis-aligned splits yield straightforward descriptions for each partition
element, recoverable by tracing a path from the root to the leaf. Finally,
predictions remain constant within each element, changing only across partition
boundaries.

\subsubsection{Advances}

Sparse and tree-based models are the foundation for many proposals for
interpretable machine learning \citep{Xin2022-ie, Zhong2023-zi}. For example, \cite{Zeng2016} introduced fast
decision trees that match black box models in recidivism
prediction, a domain with significant moral ramifications where interpretability
is crucial.  Similarly, \cite{10.5555/3524938.3525325} designed a decision tree
variant that competes effectively with deep learning in various operations
research tasks. By leveraging advances in combinatorial optimization, their
approach allows interpretable trees to be learned from very large datasets
without excessive compute demands.

Similarly, $\ell_{1}$-regularization is central to many modern interpretability
workflows, even in problems that initially seem distant from traditional
statistical learning frameworks. For example, \cite{FridovichKeil2022} developed
an interpretable approach to the view synthesis problem, where many
two-dimensional views of an object are consolidated into a coherent
three-dimensional reconstruction. Instead of the typical deep encoder-decoder
architecture, they used $\ell_{1}$-regularized regression on the appropriate
Fourier representation. This approach achieved better accuracy than competing
deep learning approaches with far less sensitivity to hyperparameter selection,
and could be trained at a fraction of the compute cost.

\subsubsection{Application to the simulation}
\label{sec:simulation_application}

We next apply these intrinsically interpretable models to the simulated trajectories
from Section \ref{subsec:simulation_design}. We explore two approaches. First,
we concatenate time points for each participant into a single $T \times
D$-dimensional vector. This method can uncover temporally localized predictive
signals. Second, we derive time series summaries for each taxon, reducing the
number of predictors to $S \times D$ for $S$ different summaries. In our study,
we focus solely on linear trends and curvature. For each subject $i$ and taxon
$d$, we fit a linear model $x_{itd} \sim \mu_{id} + \beta_{id} t$, with time as
a predictor. The estimated $\hat{\beta}_{id}$ summarizes the linear trend for
that trajectory. Similarly, we use the sum-of-squares of the temporal second
differences to summarize curvature: $\frac{1}{T - 2} \sum_{t = 2}^{T -
1}\left(x_{itd + 1} - x_{itd - 1}\right)^2$.  Note that summarization derives
potentially relevant features that are not simple linear combinations of the
original inputs. 

The holdout accuracy for the sparse logistic regression and decision tree models
under these two representations of $\*x_{i}$ and across a range of sample sizes $N$ is given in Table \ref{tab:tab1}.
Regularization $\lambda$ and tree complexity parameters were chosen with four-fold cross validation. For these data, the sparse logistic regression model
generally outperforms the decision tree. Moreover, manually derived summary statistics
yield better performance compared to concatenating all inputs. This suggests a
general takeaway: a simple model on the appropriate representation can achieve
good performance, and performance is contingent on the representation. The
widely invoked ``interpretability\textendash accuracy tradeoff'' overlooks this nuance.
Further, Table \ref{tab:tab1} includes the training time for the models,
illustrating a broad range of runtimes among interpretable models. Simple
doesn't always mean fast or scalable.

With respect to interpretability, when $N = 500$, the sparse logistic
regression model on concatenated inputs with the optimal $\lambda$ used 37 taxon-by-time point
features, while the decision tree needed 13 splits. In contrast, with the
summary statistic featurization, the optimal sparse logistic regression model
used 50 features -- the problem is lower dimensional, but the signal is more
enriched, a fact suggested by the difference in prediction performance.
Surprisingly, this enriched signal is not accessible to the decision tree, which
produces a tree with only two splits. These conclusions appear to hold across $N$. For example, even when $N = 10000$, the sparse logistic regressions on raw and featurized inputs selected 60 and 68 features, respectively.

\begin{table}[]
  \centering
  \begin{tabular}{|p{1.4cm}p{3.5cm}R{1.8cm}R{2cm}R{3cm}R{1.8cm}|}
  \hline
  \textbf{Data} & \textbf{Model} & \textbf{Samples ($\times 1000$)}  & \textbf{In-Sample Accuracy} & \textbf{Out-of-Sample Accuracy} & \textbf{Training Time (s)}\\ \hline
  Original & Sparse Logistic Reg. & 0.5  & 81.2\%         &   78.0\% &  3.9    \\
  &  & 1  & 82.1\%         &   81.0\% &  15.9   \\
  &  & 5  & 78.1\%         &   77.9\% &  22.9    \\
  &  & 10  & 77.5\%     & 77.0\%  &   37.83   \\
  & Decision Tree & 0.5 & 91.6\%        & 70.2\%  & 95.9         \\
  & & 1 & 91.3\% & 76.5\%  & 497.9         \\
  & & 5 & 91.4\% & 75.7\%  &    3255.3   \\
  & & 10 & 90.3\% & 78.4\%  & 7676.9    \\
  & Transformer    &  0.5 & 100\% & 86.4\%  & 91.2  \\ 
  & &  1 & 90.7\%  &  76.0\%  & 246.6 \\ 
  & &  5 & 96.6\% & 86.4\%  & 1167.6 \\ 
  & &  10 & 87.6\% &   87.7\%  & 1964.4 \\ 
  & Concept Bottleneck & 0.5 & 98.4\%  &  84.8\% &  110.9   \\
  & & 1 & 99.2\%   & 73.2\% &  211.6  \\
  & & 5 & 90.4\%  &   88.2\% & 892.8 \\
  & & 10 & 92.5\%   &  89.0\% &  1854.0   \\ \hline
  Featurized & Sparse Logistic Reg. & 0.5 & 92.6\%  & 87.8\% &  0.4  \\
  &  & 1 & 88.8\% &   85.4\% &  8.1  \\
  &  & 5 & 86.5\% &   85.3\% &  32.7  \\
  &  & 10 & 85.9\% &   85.5\% &  119.5 \\
  & Decision Tree & 0.5 & 74.4\%     &  69.6\%  &  1.8 \\
  & & 1 & 87.9\% &  70.6\%  &  14.7 \\
  & & 5 & 90.2\%  &  75.1\%  &  185.5 \\
  & & 10 & 88.1\%     &  77.7\%  &  770.8 \\
  \hline
  \end{tabular}
  \caption{In and out-of-sample accuracies of models applied to the simulation
  study. For cross-validated models, training time was averaged across folds.
  Training time was computed on a 2022 MacBook Pro with 16GB RAM and an M2 GPU.
  Models differ in their inherent interpretability, manual feature curation
  effort, and generalization performance. Deep learning models can overfit the
  training sample while still generalizing well. Intrinsically interpretable models
  can be substantially improved through effective featurization.}
  \label{tab:tab1}
\end{table}

\begin{figure}
 \includegraphics[width=\textwidth]{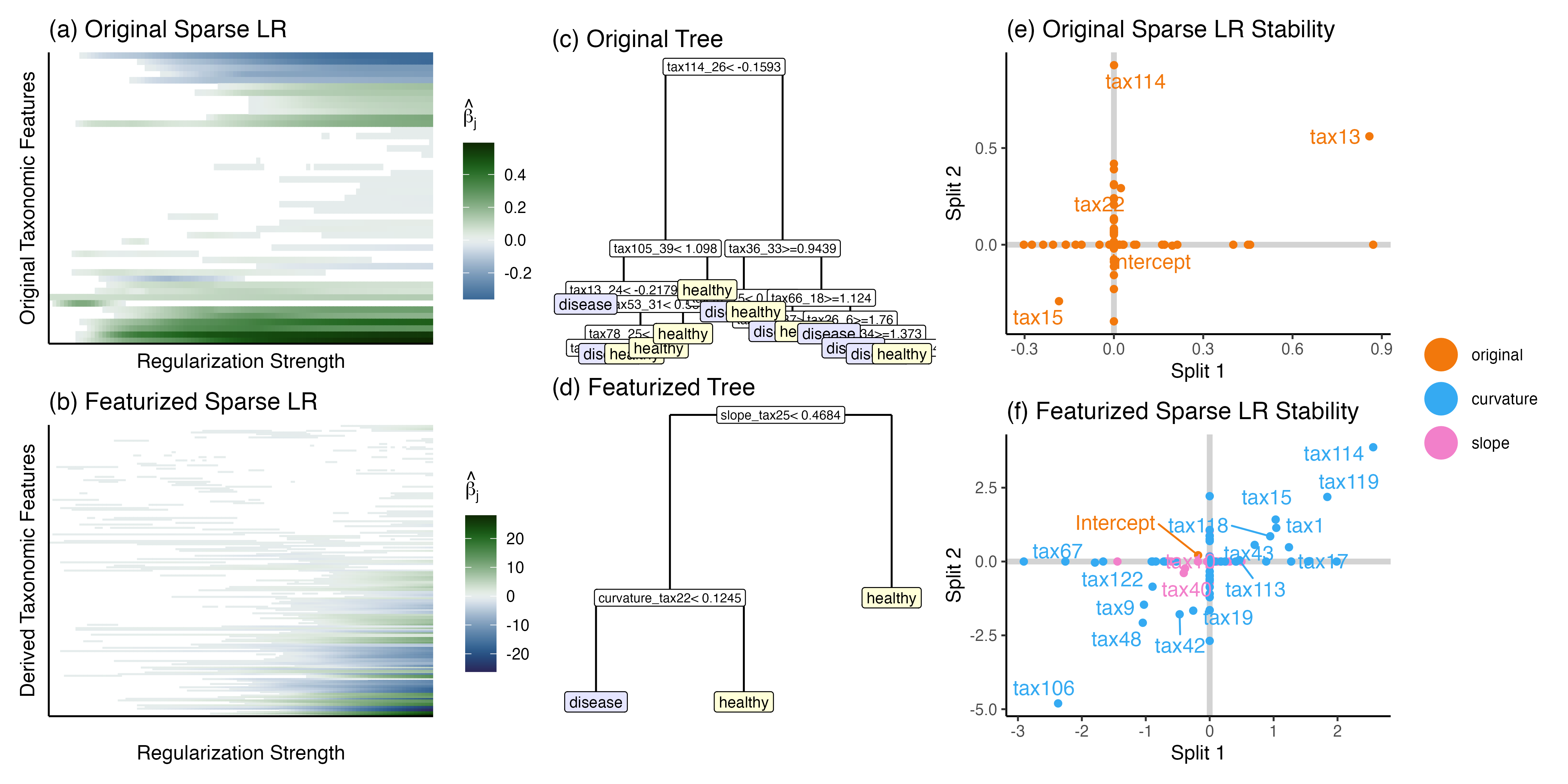}
 \caption{Results from intrinsically interpretable models applied to the simulation
 study when $N = 500$. (a) Coefficient paths on the original, unfeaturized data.
 $\ell_{1}$-regularization strength decreases from left to right. Only those
 that are nonzero for at least one penalty $\lambda$ are displayed. (b) The
 corresponding coefficient paths for the featurized data. Though there are fewer
 input features overall, more are selected. (c - d) Selected decision trees on the
 original and the featurized data, respectively. With derived features, a very
 simple tree achieves surprisingly good performance (compare with Table
 \ref{tab:tab1}) (e - f) Stability of cross-validation optimized sparse logistic regression fits on
 the original and featurized data, respectively. The regression using manually
 derived features is more stable, which is consistent with the high temporal
 correlation in the original data (see Fig \ref{fig:example_trajectories}a).}
 \label{fig:direct_interpretability}
\end{figure}

The parsimony of these estimates aligns with our discussion of the
characteristics of intrinsically interpretable models. In both approaches, prediction
relies only on a small subset of the original inputs. But how much can we trust
these features? Fig \ref{fig:direct_interpretability}e - f shows the estimated
$\hat{\beta}$ for nonzero features obtained from separately fitting the sparse
logistic regression approaches on independent splits.  When using concatenated
inputs, only two features overlap, while with summary statistics, this increases
to 18. Therefore, while the model on raw time points may be more parsimonious, it
is less stable. In this sense, despite its parsimony, the first model does not
lend itself naturally to interpretation.  Conversely, the second appears more
accurate and stable.  Moreover, its features, like curvature and trends, match
the existing vocabulary of domain experts, making it the more interpretable model
in this context. The larger lesson is that interpretability refers to a
constellation of model properties -- parsimony and simulatability but also
accuracy, stability, and contextual relevance
-- and it is worthwhile to think in these more refined terms rather than simply
``interpretable'' and ``black box'' models.

\subsection{Deep learning and XAI}

The sparse logistic regression with derived summary statistics seems to be our
best option in the case study. However, we should be concerned by the fact that
we manually define our features -- what other potentially predictive features
might we be missing out on? Black box models are appealing because they
automate the representation learning process \citep{8187120, Bengio2013}. For
example, in computer vision, deep learning models can identify higher-order,
semantically meaningful features from raw pixel inputs
\citep{Zeiler2013VisualizingAU, Simonyan14a, Yosinski2015UnderstandingNN, Achtibat2023-tg, Bykov2023-ri}. The
can learn features that activate on images of eyes simply by being given
training data of faces and an appropriate loss function \citep{Ross2017-ey}. Applying a deep learning
model to our microbiome data could reveal predictive temporal and taxonomic
features (e.g., the presence of particular subcommunities) that we had not
previously considered. The challenge, then, would be to explain these models in
a way that advances scientific knowledge and ensures accurate medical diagnoses.
To demonstrate the process, let's explore using a transformer encoder model in
our simulation.

\subsubsection{Transformers}
\label{sec:transformers}

We first review the mechanics of transformer architectures in deep learning.
Beyond our case study, transformers are particularly relevant to XAI because
they have become a universal building block of modern AI. What used to be
multiple specialized architectures -- like CNNs for vision, LSTMs for text, and
autoregressive networks for audio -- have now become consolidated into different
forms of transformers. At a high-level, the transformer is an architecture for
deriving features from sequential data. It works by recombining representations
of individual sequence elements through a series of ``attention'' operations, 
which prioritize segments of the sequence relevant to the current element's
updated representation. It takes low-level representations of individual
sequence elements, like one-hot encodings of individual words, and then modifies
them to reflect larger sequential context. 

Specifically, consider a sequence of tokens $\*x_{1}, \dots, \*x_{T}$. These tokens
can be embedded into a sequence $\*z_{1}, \dots, \*z_{T}$ of high-dimensional (e.g.,
512) real-valued vectors. In a word-based language model, $\*x_{t}$ could be
one-hot encodings of the words from the language's vocabulary \citep{Pennington2014}, while in a vision
problem, $\*x_{t}$ might represent neighboring patches from an image \citep{50650}. The $\*z_{t}$
vectors are meant to capture the high-level features of the original $t^{th}$
token (word or patch) within its larger context (sentence or image).  Therefore,
they are much more valuable for downstream analysis than word identity or pixel
values considered in isolation.

Transformer models use a mechanism called self-attention to \textit{transform}
the original $\*x_{t}$ into the $\*z_{t}$ \citep{10.5555/3295222.3295349,
lee_transformers_nodate}. If we stack the $\*x_{t}$ into $\*X \in \reals^{T \times
D}$, then self-attention has the form,
\begin{align*}
  \frac{1}{\sqrt{D_{A}}} \left(\*X \*W_{q}\*W_{k}^{T}\*X^{T}\right) \*X\*W_{v},
\end{align*}
for trainable parameters $\*W_{q} \in \reals^{D \times D_{A}}, \*W_{k} \in
\reals^{D \times D_{A}}, \*W_{v} \in \reals^{D \times D_{z}}$. The term in
parenthesis has dimension $T \times T$ and summarizes the extent to which
coordinates $t'$ in the output ``attend'' to coordinates $t$ in the input. The
matrix $\*W_{v}$ allows attention to return a linearly transformed version of
$\*X$, which can often be helpful for increasing/decreasing output dimensions.
Typically, several of these transformations are applied in parallel,
concatenated into an object called ``multihead attention,'' and passed through
a softmax nonlinearity. The output of this complete process is the matrix of
embeddings with rows $\*z_{1}, \dots, \*z_{T}$.

The embeddings $\*z_{t}$ are difficult to interpret because their derivation is
neither parsimonious nor simulatable. Empirically, the representations have been
shown to encapsulate high-level features associated with each token's context.
However, the number of updates, the fact that the update can ``attend'' to all
other tokens simultaneously, and the fact that the attention mechanism is
nonlinear all contribute to the difficulty in interpreting these embeddings as
well as any decision rules derived from them. Nonetheless, several approaches
have emerged for enhancing the interpretability of these types of deep,
embedding-based models \citep{JMLR:v11:erhan10a,
Wongsuphasawat2018VisualizingDG}. We review three approaches below: global
explainability, integrated gradients, and concept bottleneck models.

\subsubsection{Embedding visualization}
\label{sec:embedding_visualization}

Embedding visualizations treat the learned representations $\*z_{t}$ as data and
apply Exploratory Data Analysis techniques to better understand their
properties. Methods like principal components analysis, canonical correlation
analysis, or UMAP are often applied to the collection of $\*z_{i}$
\citep{Nguyen2019, raghu_vision_2021, 10.5555/3454287.3454588}. Since the
embeddings can be studied simultaneously across the full dataset, this is an
example of a global explainability technique. 

For example, \citet{10.5555/3454287.3455058} used embeddings to analyze the BERT
language model \citep{devlin-etal-2019-bert}. In one experiment, they examined
sentences containing words that can have multiple meanings depending on context,
like, ``fair.'' By computing embeddings for instances of ``fair'' within these
sentences, it was observed that although the word remained the same, its
embeddings distinctly clustered in a two-dimensional principal components
projection. This clustering reflected the context dependence of the word's
meaning, with clusters representing the legal (``fair use''), mathematical
(``fair coin''), and civic (``world's fair'') uses. This qualitative
study can also be accompanied by quantitative analysis. For example,
\cite{10.5555/3454287.3455058} and \cite{hewitt-manning-2019-structural} used
linear probes to quantify the extent to which grammatically structure
(represented by parse tree representations) could be reconstructed from BERT
embeddings. 
Figure \ref{fig:embeddings_combined} in Section \ref{sec:simulation_application}
gives embedding visualization examples in the context of our simulation study. In particular, Figure \ref{fig:embeddings_combined}d parallels the approach of \citet{10.5555/3454287.3455058}, focusing on a species instead of a word.

\subsubsection{Integrated gradients}

When seeking local explanations, embedding visualization lacks sufficient
specificity. We often have questions in mind that require more than model-based
summaries -- for example, what was it about a hotel review that led to its
classification as negative sentiment, or how could a loan applicant modify their
application so that it can be approved? Addressing these questions requires
thinking carefully about the model's sensitivity to perturbation. Many methods
formalize this intuition -- influence functions \citep{10.5555/3305381.3305576},
Shapely values \citep{10.5555/3295222.3295230, Fumagalli2023-vl}, LIME
\citep{10.1145/2939672.2939778}, GradCAM \citep{8237336}, and Integrated
Gradients (IG) \citep{10.5555/3305890.3306024} for example. Here, we dive into
the mechanics of IG.

The integrated gradient of a sample $\*x_{i}$ with respect to class $y$ defined by:
\begin{align}
  \text{IG}\left(\*x_{i}\right) &= \left(\*x_i - \*x_0\right) \times \int_{\alpha = 0}^{1} \frac{\partial f_{y}\left(\*x_{0} + \alpha \times \left(\*x_{i} - \*x_{0}\right)\right)}{\partial \*x_{i}} d\alpha,
\end{align}
where $\*x_{0}$ is a reference sample (e.g., an all-black ``empty'' image) and
$f_{y}$ is the model's predicted probability for the input being assigned to
class $y$. The gradient $\frac{\partial f_{y}\left(\*u\right)}{\partial
\*u}$ represents how infinitesimal
changes in each coordinate of $\*u$ affects the probability of assigning
the input to class $y$. At first, this appears to be sufficient
for local explanation -- large gradient coordinates correspond to words or pixels
with the largest influence on the class assignment. However, this often fails
in practice due to saturation. Specifically, deep learning
classifiers often use a multiclass logistic function in the final layer, and for
large input magnitudes, this function's gradients approach zero, rendering
explanations based on $\frac{\partial f_{y}\left(\*u\right)}{\partial \*u}$
unsatisfactory.

The solution proposed by IG is to consider versions of the input $\*x_{0} +
\alpha \times \left(\*x_{i} - \*x_{0}\right)$ that have been shrunk down to $\*x_0$, hence
avoiding vanishing gradients. Since it might not be clear how much to shrink at
first, IG considers scaling along an entire sequence of $\alpha \in \left[0,
1\right]$. Integrating gradients over this range gives a more faithful
description of the $i^{th}$ coordinate's contribution to classification as $y$.
For example, in \cite{Sturmfels2020}, gradients for the object of interest (a
bird) are much larger for small $\alpha$ values, ensuring that the final IG for
the bird class places appropriate weight on that class.

Integrated gradients and its local explanation counterparts are widely used in
XAI, but they are also among the most controversial \citep{pmlr-v162-lundstrom22a}.  Both theoretical and
empirical results suggest that explanations do not necessarily respect
expectations. For example, \cite{10.5555/3327546.3327621} observed that saliency
maps, including those from IG, derived from randomly initialized neural network
classifiers look closely resemble those from properly trained ones. A
theoretical study by \cite{Bilodeau2024} found that locally similar functions 
could nonetheless be assigned different explanations at their query points,
indicating that global properties can influence local explanations. These
alarming findings have renewed interest in the interpretability community to
develop local explainability techniques with formal guarantees.

\subsubsection{Concept bottleneck models}
\label{sec:concept_bottleneck_models}
The explainability techniques we have discussed so far depend only on the model and either the original training data, for embedding analysis, or the specific example to be explained, for integrated gradients.
This is both a strength and a
limitation. On the one hand, it ensures generality, the same explanation method
can be effective across various applications. On the other, it limits the
incorporation of domain-specific language into the explanation. Concept
Bottleneck Models (CBMs) \citep{50351} offer an abstraction for infusing
domain-specific language into models.  CBM ``concepts'' bridge model and human
representations of relevant data features. The overarching goal of this class of
models is to create representations that can be manipulated by both the model
(as high-dimensional, numerical vectors) and experts (as community-derived
concepts).

CBMs leverage expert knowledge to form annotations useful for both model
training and explanation. During training, each sample has the form
$\left(\*x_{i}, y_{i}, \*c_{i}\right)$. $\*x_{i}$ and $y_{i}$ are the usual supervised
model input and output for sample $i$. $\*c_{i} \in \{0, 1\}^{K}$ represents a
dense concept annotation. For example, $\*x_{i}$ could be the image of a bird,
$y_i$ its species label, and $\*c_i$ a summary of interpretable features like wing
color or beak length.  The CBM learns a model $\*x_i \to \*c_i \to y_i$, compressing
low-level measurements in $\*x_i$ into concept annotations $\*c_i$ before making a
prediction $y_i$. \cite{50351} propose several strategies for implementing this
bottleneck, including joint training of functions $f$ and $g$ to minimize the
weighted sum of concept $L_{1}$ and label $L_{2}$ losses:
\begin{align}
\hat{f}, \hat{g} := \arg\min_{f,g} \sum_{i} L_{1}\left(\*c_i, f\left(\*x_i\right)\right) + \lambda L_{2}\left(y_i, g\left(\*c_i\right)\right),
\end{align}
for some $\lambda > 0$.

On a test sample $\*x^\ast$, predictions are made using $\hat{\*c} =
f\left(\*x^{\ast}\right)$ and $\hat{y} = g\left(\hat{\*c}\right) =
g\left(f\left(\*x^\ast\right)\right)$. Note that ground truth concept labels $\*c_{i}$ are not
needed at this stage. Typically, a simple model $g$, like multiclass logistic
regression, is used. CBMs automate human-interpretable
feature extraction, bridging manual feature engineering with black box
representation learning. While manually designing a ``wing color'' feature
extractor would be tedious, it is helpful to know that the model $f$ learns a
proxy for such a feature, and its output can be used like any human engineered
feature. Moreover, this approach streamlines counterfactual reasoning. For
example, we can ask what the model would have predicted had the wing been yellow
instead of red, simply by intervening on the coordinate of $\*c_{i}$ associated with wing color.
This is important in fairness or recourse analysis \citep{Mehrabi2021, Karimi2022}. For example, if we know that
a feature should not be used for prediction (e.g., the race of a loan
applicant), then we can zero out the coefficient from that concept in the model
$g$.

\begin{lstlisting}[language=Python,caption={An example implementation of a Concept Bottleneck Model. The variable \texttt{c} represents the concept labels, which are transformed into class predictions using a shallow multilayer perceptron, \texttt{self.mlp}. See the full implementation at the repository \href{https://go.wisc.edu/3k1ewe}{https://go.wisc.edu/3k1ewe}.}, label={lst:concept}]
class ConceptBottleneck(nn.Module):
 """
 Learn Concepts and Classes for Sequences
 """
 def __init__(self, n_embd=144, n_positions=50, n_layer=6, n_concept=25, n_class=2):
  super(ConceptBottleneck, self).__init__()
  self.n_concept = n_concept
  self.n_class = n_class
  config = GPT2Config(n_embd=n_embd, n_positions=n_positions, n_layer=n_layer)
  self.backbone = GPT2Model(config)
  self.concept = nn.Linear(n_embd * n_positions, self.n_concept)
  self.mlp = nn.Sequential(
    nn.Linear(self.n_concept, self.n_concept),
    nn.ReLU(),
    nn.Linear(self.n_concept, self.n_concept),
    nn.ReLU(),
    nn.Linear(self.n_concept, self.n_concept),
    nn.ReLU(),
    nn.Linear(self.n_concept, self.n_class - 1)
  )

 def forward(self, x):
  z = self.backbone(inputs_embeds=x)
  c = self.concept(z.last_hidden_state.view(x.shape[0], -1))
  return c, self.mlp(c)
\end{lstlisting}

\subsubsection{Application to the simulation}
With these XAI techniques, we can apply black box models to our microbiome case
study and still draw scientific insight from them. The simulated dataset is
well-suited to a transformer encoder architecture, where each sample corresponds
to a word and the full trajectory to a sentence.  The final healthy vs. disease
classification is analogous to text sentiment classification.  Guided by these
parallels, for each $N$ we trained a GPT-2 architecture with a random sample of $0.75 N$ simulated
participants; the remaining $0.25N$ are reserved for validation. We used default
hyperparameters from a public implementation with the two modifications. First,
we changed the input dimension to 144 to one-hot encode the species. Second, we
reduced the number of layers to 6 from the default of 12, since the sample sizes
are relatively small. After training this model for 70 epochs, the model achieves
a holdout accuracy ranging from 76.0\% ($N = 1000$) to 87.7\% ($N = 10000$). This performance is noteworthy considering that 
in several cases it exceeds the best models with handcrafted features (cf. Table
\ref{tab:tab1}), which were designed with the true generative mechanism in mind.
Indeed, considering the close alignment between handcrafted features and the
generative mechanism, we suspect that the Bayes error for this problem is between 10 and 15\%.
The ability to match or exceed a handcrafted model underscores the appeal
of modern deep learning. It is not necessarily about achieving better
performance than interpretable counterparts on all problems;
rather, it comes from the potential to bypasses feature engineering -- a
traditionally labor-intensive and time-consuming step -- while achieving
comparable performance. Moreover, this simulation highlights the attractive scaling of deep learning models in both out-of-sample accuracy and running time, since performance continues to improve with larger sample sizes. This makes them valuable in applications where frequent errors cannot be tolerated, even if they could be explained. This is often the case in technological systems that have to operate well even with limited human oversight, like flight control \citep{Chai2021} or online recommender systems \citep{resnick1997recommender}. Indeed, in these settings, reliability and robustness may be more important than interpretability alone.

\begin{figure}
\includegraphics[width=\textwidth]{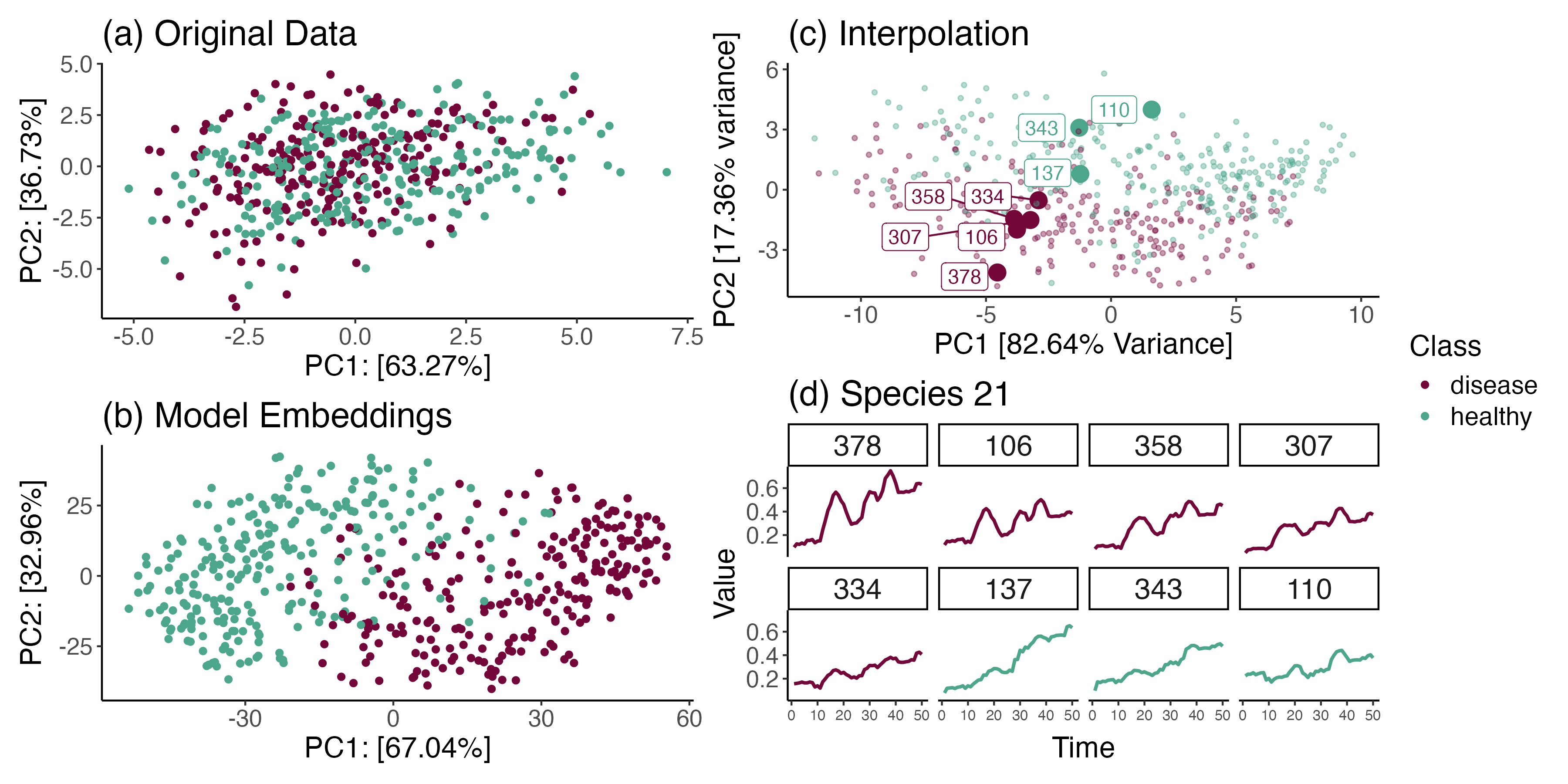}
\caption{Global explanation of the simulation's transformer model via
embeddings. (a) Healthy and disease classes generally overlap in the original
data. Each point gives the projection of a subject onto the top two sparse PCA
directions. (b) Transformer representations more clearly separate the two
classes. (c) A sequence of points along the linear interpolation between samples
110 and 378 in embedding space restricted to features from species 21.  (d)
Trajectories associated with each labeled point in (c).}
\label{fig:embeddings_combined}
\end{figure}

The remaining challenge lies in understanding how the models achieved their
performances. We consider each of the techniques discussed above, focusing on the models trained with $N = 500$ samples. Fig
\ref{fig:embeddings_combined} offers two global views of the data. Fig
\ref{fig:embeddings_combined}a gives the sparse PCA projections derived from
the scaled $N \text{ subjects} \times \left(D \text{ taxa} \times T \text{
timepoints}\right)$ matrix of community profiles over time, while Fig
\ref{fig:embeddings_combined}b shows the analog for the $N \text{ subjects}
\times L \text{ embeddings}$ transformer representations. The embeddings in (b)
more clearly distinguish healthy and diseased groups, capturing more of the
total variance within the top two dimensions. This is expected, since the
final classification uses a linear hyperplane, encouraging the transformer to
learn a linearly separable representation. To better understand
the features encoded by the transformer, we focus on the subset of embeddings
associated with species 21. Fig \ref{fig:embeddings_combined}c shows sparse PCA
applied to these embeddings. We have highlighted samples that lie near
the linear interpolation between samples 110 and 378 in the original embedding
space; their original trajectories are shown in panel (d). Along this
interpolation, the peaks and valleys become less pronounced, suggesting that the
model has learned to differentiate curves with different trajectory properties. This is a
subtle feature, and one that we know by design is related to the response. In
contrast, some visually prominent features unrelated to disease, like the
maximum value of the series, are appropriately disregarded.

\begin{figure}
\includegraphics[width=\textwidth]{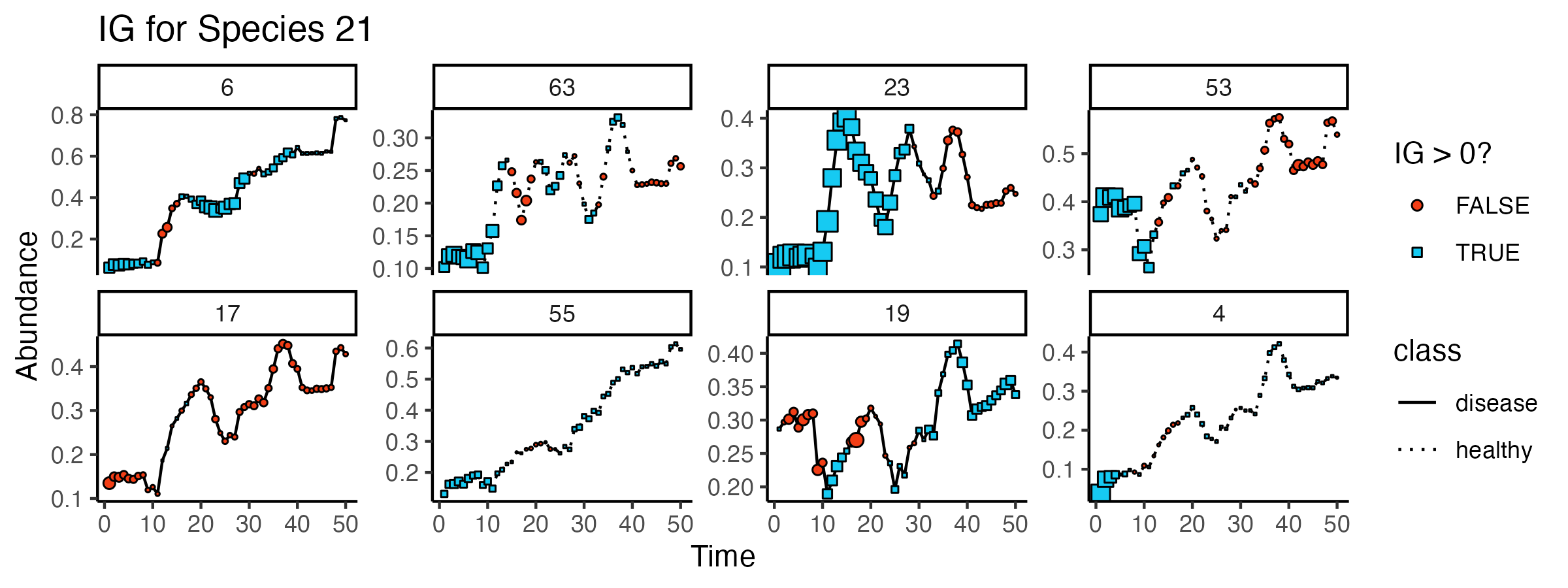}
\caption{Integrated gradients for the transformer model overlaid on a subset of
samples from the simulation study. Each curve represents the trajectory for
species 21, with symbols indicating the size and sign of the integrated
gradient at that sample and time point. The subset of relevant time points can
vary from sample to sample.}
\label{fig:integrated_gradients}
\end{figure}

We next study local explanations using IG. Based on the input structure, IG
returns a value for each time point-by-taxon combination. Example values are
given in Fig \ref{fig:integrated_gradients}. In each panel, we can identify
time points that informed the final classification and the direction in which
perturbations change the target class label. Blue squares denote that an increased
abundance at that time point increases the probability of correct classification;
red circles suggests the opposite. Larger symbols imply larger changes in probability.
For example, for subject 6, the first ten time points and the valley surrounding
time 20 were most important. Increases in either of these intervals increase the
probability of correctness. In contrast, for subject 63, an increase in the
valley surrounding time 20 decreases this probability. Since these subjects
are from opposite classes, deeper valleys around time 20 seem to be related to
health. However, beware that these conclusions are essentially local, and this
pattern need not hold across all samples for IG to declare it important. In
general, IG effectively localizes the responsibility for a classified instance
down to a subset of time points, aiding error analysis and uncovering subsets
of features that could flip the class of a prediction.

Next, we take a closer look at the use of CBMs in our case study. Recall that
concepts are a form of dense annotation for each sample. We extend our original
generative mechanism to provide sample-level concepts. For each
$\theta_{ik}$, we define a binary vector by thresholding: $\*c_{i} =
\mathbf{1}\{\theta_{ik} > t\}$. Since each $\theta_{ik}$ corresponds to a
latent trajectory, this acts as a coarse label for sample $i$,
indicating which of the $k$ underlying trajectories it is most influenced by.
Next, we adapt our transformer to predict both the binary pattern of $\*c_{i}$
and the final disease status, adding a concept-level loss that computes a
logistic loss coordinatewise (see Listing \ref{lst:concept}). The results are
displayed in Table \ref{tab:tab1}. Notably, using this concept definition, the
model performs better than the original, unconstrained transformer, suggesting
closer alignment to the original generative mechanism. In general, better CBMs
require denser concept annotation. Note, however, that recent work has proposed
variations that discover concepts de novo or identify cheap concept proxies
following model training \citep{yuksekgonul2023posthoc,
10.5555/3454287.3455119}.

\begin{figure}
 \includegraphics[width=\textwidth]{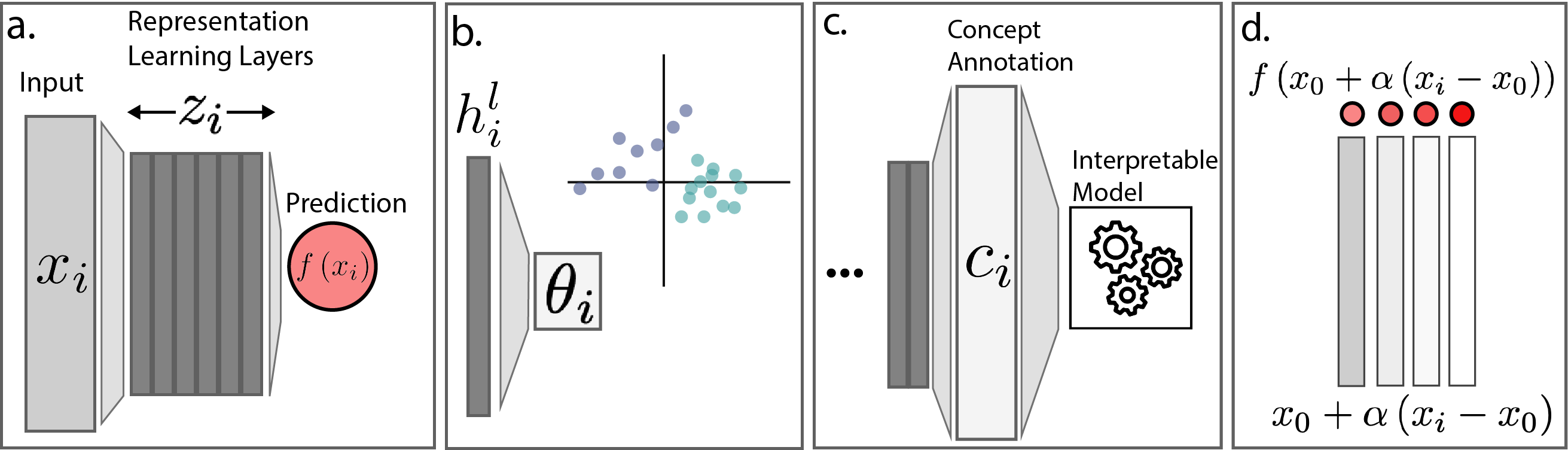}
 \caption{Summary of XAI techniques. (a) Deep learning models transform input
 $\*x_{i}$ into higher-level semantic representations. The final layer is used for
 prediction. (b) Global embeddings represent a layer of interest with
 lower-dimensional vectors $\*z_{i}$, which can be interpreted alongside
 sample-level metadata, like point color in this example. (c) CBMs only use the
 black box to predict dense, concept-level annotations, not the final response.
 An interpretable model translates predicted concepts into the prediction (d) In
 IG, a sample is subject to a sequence of perturbations, and the impact on the
 final prediction is recorded to identify important coordinates.}
 \label{fig:xai-summary}
\end{figure}

\section{Evaluation}
\label{sec:evaluation}

Developing effective benchmarks for interpretability methods is an active area
of research, critical for objectively comparing proposals.  However, designing
effective criteria faces several challenges,

\begin{itemize}
\item Operationalization: Interpretability encompasses several technical
definitions, like parsimony and simulatability, making it difficult to capture
precisely with a single metric \citep{JMLR:v24:22-0142, Hedstrom2023-gd}.
\item Output Types: Different classes of methods return different types of
outputs, like attribution maps and concepts, which are hard to compare directly.
\item Context Dependence: Method performance must be gauged relative to the
audience and tasks they were designed for. Nonetheless, a good benchmark should
generalize across tasks.
\item Ambiguous Ground Truth: Unlike supervised learning, establishing ground truth
interpretability output (e.g., the ``correct'' feature attribution map) is often
impossible \citep{Arras2022-oq, Guidotti2021-bv}.
\item Human-Computer Interaction: The success of an interpretability method
depends on how it interfaces with users \citep{JMLR:v22:20-1473, Baniecki2019}. Simplified experiments may not reflect
real-world complexity, while real-world observations may be confounded by
factors that don’t generalize across applications.
\end{itemize}

These challenges mirror those encountered in data visualization, which has
developed various evaluation protocols to guide systematic progress. Unlike
traditional machine learning, where isolated computational benchmarks have led
to significant progress, we expect evaluation in interpretability will depend on
triangulating multiple sources of experimental evidence, like in data
visualization. 

Interpretability evaluation methods fall into two general categories: tailored
benchmark datasets and user studies with controlled tasks. Within each class,
there are variations that account for the challenges above in different ways.
For dataset benchmarks, one approach is ablation \citep{hooker_benchmark_2019,
Wang2022-aj}.  Specifically, for methods that output feature importance
attributions, this involves retraining models with versions of the data where
purported important regions are masked out. Deterioration in model performance
suggests that the attributions accurately reflect regions that the original
model relied upon.  Note that retraining is necessary. The ablated images are
not drawn from the same distribution as the training data, so deterioration in
performance of the original model could be due to either removal of important
features (the criteria of interest) or domain shift (a confounding factor).
Another data-driven benchmarking approach is to design synthetic datasets. These
datasets are generated using known mechanisms, allowing for later evaluation of
explanations \citep{Liu2021-ao, Zhou2022-lb}. For example, the funnybirds benchmark creates imaginary cartoon
birds where ``species'' classes are derived from attributes, like wing, beak,
and tail type, that are controlled during data generation \citep{Hesse2023-zq}.
Given a specific instance, an ideal explanation should focus on the parts of the
bird that distinguish it from related classes.  To make these benchmarks more
realistic, variation in viewpoint and illumination can be introduced.

In user studies, one approach is to have participants make pairwise preference
judgments between explanations \citep{jeyakumar_how_2020}. However, subjective
preference may not align with objective task performance. To this end, more
specific experimental tasks can be designed, evaluating user decisions with and
without support from interpretability outputs \citep{Ma2024-nt,
colin_what_2022}. For example, to assess simulatability, participants could be
asked to guess a model’s output on an example before and after viewing an
explanation.  Alternatively, they could be asked to guess how predictions
might change under perturbations to the input -- effective explanations
should help in making accurate guesses. Finally, integrating interpretability
methods into workflow-specific interfaces makes it possible to study their
performance within tailored tasks. For example, \cite{Wu2021} designed an
interface to aid auditors in generating counterfactual examples that flip model
classifications. This is an instance of a counterfactual explanation \citep{Mothilal2020-zz, Guidotti2022-zv} that helps to characterize sensitivity to out-of-distribution
shifts. Influential examples that participants discovered could be incorporated
into training to improve generalization. Different choices in the frontend
interface and backend interpretability engine led to differential success in
discovering surprising counterfactuals (those with small perturbations in input
that led to large change in class predictions). In this way, interface-oriented
user studies can more precisely gauge ``human-AI collaboration.'' Indeed, these
experiments have uncovered counterintuitive phenomena, like instances where task
performance worsens with AI explanations — human ``collaborators'' can become
less critical when shown an explanation \citep{Bansal2021}. Finally, a key
challenge in user studies is selecting an appropriate participant pool with
representative tasks \citep{Buinca2020}.  For example, an interpretability
method in healthcare may be more effective for patients than for doctors. Narrow
tasks can more directly guide deployment while broader ones are easier to run at
scale.

There is often overlap between benchmark datasets and user study evaluation. For
example, \cite{casper_red_2023} created a dataset to understand the
effectiveness of explanations in aiding model debugging. The dataset introduced
distractor patches to a small fraction of images, leading to failures on the
test set. For example, a small strawberry was pasted into 1 in 3000 images in
both training and test sets, and in each ``poisoned'' training (but not test) image, the class
was switched to ``goose,'' leading to failures at test time.  Participants were
presented with misclassified test images, along with various explanations.
Their accuracy and speed in identifying the poisoning mechanism served to
quantify the utility of these explanations for debugging.

Evidently, while there is no one way to evaluate interpretability methods, there
is a natural hierarchy of tests to which proposals can be subjected.
General-purpose proposals should first pass ablation and synthetic data
benchmarks. Success in these criteria justifies investing resources in user
studies with more realistic audience and task designs \citep{Rong2024-ss}. Future interpretability
methods may be evaluated similarly to new therapies in medicine: \emph{in silico}
associations (ablation and synthetic data studies) can guide \emph{in vitro}
experiments (user studies with generic audiences) which, if successful, lead to
follow-up clinical trials (user studies with representative audiences).

\section{Discussion}
\label{sec:discussion}

This review has surveyed the significant advances in interpretability over
recent decades. In closing, we highlight key ongoing shifts in the AI landscape
and their implications for interpretability. One important trend is the rise of
multimodal models, which are trained on mixed data modalities
\citep{10.5555/3104482.3104569, chen_history_2021}. This has facilitated novel
applications, like automatic open vocabulary segmentation
\citep{zou_segment_2023}, where, rather than annotating each image pixel with
one of a preset collection of class labels, the system can generate segmentation
masks for arbitrary text strings (e.g., ``The pedestrian that is crossing the
street.''). While single modality methods remain applicable to individual model
components (e.g., analyzing the word embeddings in a vision language model),
there is a pressing need for interpretability techniques that offer an
integrated view of how these models merge information across sources. This could
help developers train models that effectively balance evidence from
complementary views, enhancing overall system robustness. Statistical methods
for data integration \citep{Ding2022} could offer a principled approach for
analyzing models with these more complex, interacting components.

Another significant trend is the emergence of foundation models
\citep{Bommasani2021-ie}. These models are trained on corpora spanning petabytes
of data, demonstrate a capacity for in-context learning. They can immediately
solve tasks that previously required additional labeled data and fine-tuning. In
this sense, foundation models are generalists, allowing users across diverse
domains to benefit from a single, albeit intensive, modeling investment.
Therefore, understanding how to interpret these models and their role in
downstream specialized tasks is an important area for study. It is no longer
wise to explain AI systems as if they were trained in isolation; they must be
considered part of a broader AI ecosystem. Further,
\cite{DBLP:conf/iclr/XieRL022} and \cite{teh_statistical_nodate} have pointed
out connections between in-context learning and Bayesian posterior updating.
This suggests that statistical thinking may resolve attribution questions when
data are drawn from heterogeneous sources, which is important for resolving data
provenance and ownership.

Third, as new legal and social norms around model-based decision-making are
established, we will likely encounter scenarios where even comprehensive
explanation are unsatisfactory, and where high-performing glass boxes become
critical \citep{Nannini2023-ep}. In certain applications, reliability and interpretability will
outweigh performance alone. For example, in healthcare and legal contexts,
simple decision-making criteria are ideal, and in science, generalizable insight
holds greater value than out-of-sample performance alone. Though recent
progress in deep learning has sparked the community's imagination about
potentially more human-like or human-surpassing systems, we should perhaps
consider futures where systems only marginally outperform current ones but are
significantly safer and transparent.

Finally, we ask: how can we make interpretability a more interactive process?
Interactivity is critical for ensuring that our systems augment intelligence,
empowering users to become creative and critical problem solvers \citep{Baniecki2023-of, Sokol2020-de}. Neglecting
this problem may lead to an overreliance on oracles, even among experts. For
example, doctors should feel comfortable rejecting AI recommendations, even if
there is a chance their decision turns out to be incorrect. A prerequisite for
such interactivity is shared language  \citep{Heer2019, crabbe_explaining_2021,
medium_beyond_nodate} -- it is difficult to interact without communication. CBMs
are already a step in this direction, where concepts serve as a representation
linking both human and machine reasoning. The goal of a shared language isn't as
fantastical as it might seem; programming, after all, is an exercise in
communicating with machines, with the programming language as the shared
representation.

Looking ahead, we imagine a dynamic where interpretability is central to the
interactive design and development of AI systems. Returning to our visualization
analogy, it is already common for national newspapers to feature sophisticated
interactive visualizations, and college freshman across various disciplines are
taught data literacy and visual exploration. In the future, interpretability can
help AI models be communicated and developed in a similarly democratic way,
making it easier for data to shape human judgment and creativity.

\section*{Supplementary Material}

Code to reproduce our simulation experiment can be found at \href{https://go.wisc.edu/3k1ewe}{https://go.wisc.edu/3k1ewe}. The repository's README file describes how to generate the dataset, fit each model, and create the visualizations. The data, compiled notebooks, intermediate outputs, and log files can be accessed at \href{https://go.wisc.edu/v623lq}{https://go.wisc.edu/v623lq}.

\section*{Funding}

This work was supported in part by grant number R01GM152744 from the National Institute of General Medical Sciences.

\bibliographystyle{jds}
\bibliography{biblio}

\begin{thebibliography}{109}
\providecommand{\natexlab}[1]{#1}

\bibitem[{Achtibat et~al.(2023)Achtibat, Dreyer, Eisenbraun, Bosse, Wiegand,
  Samek et~al.}]{Achtibat2023-tg}
Achtibat R, Dreyer M, Eisenbraun I, Bosse S, Wiegand T, Samek W, et~al. (2023).
\newblock From attribution maps to human-understandable explanations through
  concept relevance propagation.
\newblock \emph{Nat. Mach. Intell.}, 5(9): 1006--1019.

\bibitem[{Adebayo et~al.(2018)Adebayo, Gilmer, Muelly, Goodfellow, Hardt, and
  Kim}]{10.5555/3327546.3327621}
Adebayo J, Gilmer J, Muelly M, Goodfellow I, Hardt M, Kim B (2018).
\newblock Sanity checks for saliency maps.
\newblock In: \emph{Proceedings of the 32nd International Conference on Neural
  Information Processing Systems}, NIPS'18, 9525–9536. Curran Associates
  Inc., Red Hook, NY, USA.

\bibitem[{Agrawala et~al.(2011)Agrawala, Li, and
  Berthouzoz}]{10.1145/1924421.1924439}
Agrawala M, Li W, Berthouzoz F (2011).
\newblock Design principles for visual communication.
\newblock \emph{Commun. ACM}, 54(4): 60–69.

\bibitem[{Arras et~al.(2022)Arras, Osman, and Samek}]{Arras2022-oq}
Arras L, Osman A, Samek W (2022).
\newblock {CLEVR-XAI}: A benchmark dataset for the ground truth evaluation of
  neural network explanations.
\newblock \emph{Inf. Fusion}, 81: 14--40.

\bibitem[{Baniecki and Biecek(2019)}]{Baniecki2019}
Baniecki H, Biecek P (2019).
\newblock modelstudio: Interactive studio with explanations for ml predictive
  models.
\newblock \emph{Journal of Open Source Software}, 4(43): 1798.

\bibitem[{Baniecki et~al.(2021)Baniecki, Kretowicz, Piątyszek, Wiśniewski,
  and Biecek}]{JMLR:v22:20-1473}
Baniecki H, Kretowicz W, Piątyszek P, Wiśniewski J, Biecek P (2021).
\newblock dalex: Responsible machine learning with interactive explainability
  and fairness in python.
\newblock \emph{Journal of Machine Learning Research}, 22(214): 1--7.

\bibitem[{Baniecki et~al.(2023)Baniecki, Parzych, and Biecek}]{Baniecki2023-of}
Baniecki H, Parzych D, Biecek P (2023).
\newblock The grammar of interactive explanatory model analysis.
\newblock \emph{Data Min. Knowl. Discov.}, 1--37.

\bibitem[{Bansal et~al.(2021)Bansal, Wu, Zhou, Fok, Nushi, Kamar
  et~al.}]{Bansal2021}
Bansal G, Wu T, Zhou J, Fok R, Nushi B, Kamar E, et~al. (2021).
\newblock Does the whole exceed its parts? the effect of ai explanations on
  complementary team performance.
\newblock In: \emph{Proceedings of the 2021 CHI Conference on Human Factors in
  Computing Systems}, CHI ’21. ACM.

\bibitem[{Bengio(2009)}]{8187120}
Bengio Y (2009).
\newblock \emph{Learning Deep Architectures for AI}.
\newblock NOW.

\bibitem[{Bengio et~al.(2013)Bengio, Courville, and Vincent}]{Bengio2013}
Bengio Y, Courville A, Vincent P (2013).
\newblock Representation learning: A review and new perspectives.
\newblock \emph{IEEE Transactions on Pattern Analysis and Machine
  Intelligence}, 35(8): 1798–1828.

\bibitem[{Bilodeau et~al.(2024)Bilodeau, Jaques, Koh, and Kim}]{Bilodeau2024}
Bilodeau B, Jaques N, Koh PW, Kim B (2024).
\newblock Impossibility theorems for feature attribution.
\newblock \emph{Proceedings of the National Academy of Sciences}, 121(2).

\bibitem[{Bommasani et~al.(2021)Bommasani, Hudson, Adeli, Altman, Arora, von
  Arx et~al.}]{Bommasani2021-ie}
Bommasani R, Hudson DA, Adeli E, Altman R, Arora S, von Arx S, et~al. (2021).
\newblock On the opportunities and risks of foundation models.

\bibitem[{Borkin et~al.(2013)Borkin, Vo, Bylinskii, Isola, Sunkavalli, Oliva
  et~al.}]{6634103}
Borkin MA, Vo AA, Bylinskii Z, Isola P, Sunkavalli S, Oliva A, et~al. (2013).
\newblock What makes a visualization memorable?
\newblock \emph{IEEE Transactions on Visualization and Computer Graphics},
  19(12): 2306--2315.

\bibitem[{Bu\c{c}inca et~al.(2020)Bu\c{c}inca, Lin, Gajos, and
  Glassman}]{Buinca2020}
Bu\c{c}inca Z, Lin P, Gajos KZ, Glassman EL (2020).
\newblock Proxy tasks and subjective measures can be misleading in evaluating
  explainable ai systems.
\newblock In: \emph{Proceedings of the 25th International Conference on
  Intelligent User Interfaces}, IUI ’20. ACM.

\bibitem[{Bykov et~al.(2023)Bykov, Kopf, Nakajima, Kloft, and
  H{\"o}hne}]{Bykov2023-ri}
Bykov K, Kopf L, Nakajima S, Kloft M, H{\"o}hne MMC (2023).
\newblock Labeling neural representations with inverse recognition.

\bibitem[{Caruana et~al.(2015)Caruana, Lou, Gehrke, Koch, Sturm, and
  Elhadad}]{10.1145/2783258.2788613}
Caruana R, Lou Y, Gehrke J, Koch P, Sturm M, Elhadad N (2015).
\newblock Intelligible models for healthcare: Predicting pneumonia risk and
  hospital 30-day readmission.
\newblock In: \emph{Proceedings of the 21th ACM SIGKDD International Conference
  on Knowledge Discovery and Data Mining}, KDD '15, 1721–1730. Association
  for Computing Machinery, New York, NY, USA.

\bibitem[{Casper et~al.(2023)Casper, Bu, Li, Li, Zhang, Hariharan
  et~al.}]{casper_red_2023}
Casper S, Bu T, Li Y, Li J, Zhang K, Hariharan K, et~al. (2023).
\newblock Red {Teaming} {Deep} {Neural} {Networks} with {Feature} {Synthesis}
  {Tools}.
\newblock In: \emph{Advances in {Neural} {Information} {Processing} {Systems}}
  (A~Oh, T~Naumann, A~Globerson, K~Saenko, M~Hardt, S~Levine, eds.), volume~36,
  80470--80516. Curran Associates, Inc.

\bibitem[{Chai et~al.(2021)Chai, Tsourdos, Savvaris, Chai, Xia, and
  Philip~Chen}]{Chai2021}
Chai R, Tsourdos A, Savvaris A, Chai S, Xia Y, Philip~Chen C (2021).
\newblock Review of advanced guidance and control algorithms for
  space/aerospace vehicles.
\newblock \emph{Progress in Aerospace Sciences}, 122: 100696.

\bibitem[{Chen et~al.(2021)Chen, Guhur, Schmid, and Laptev}]{chen_history_2021}
Chen S, Guhur PL, Schmid C, Laptev I (2021).
\newblock History {Aware} {Multimodal} {Transformer} for
  {Vision}-and-{Language} {Navigation}.
\newblock In: \emph{Advances in {Neural} {Information} {Processing} {Systems}}
  (M~Ranzato, A~Beygelzimer, Y~Dauphin, PS~Liang, JW~Vaughan, eds.), volume~34,
  5834--5847. Curran Associates, Inc.

\bibitem[{Cleveland(1993)}]{Cleveland1993-pd}
Cleveland WS (1993).
\newblock \emph{Visualizing Data}.
\newblock Hobart Press.

\bibitem[{Coenen et~al.(2019)Coenen, Reif, Yuan, Kim, Pearce, Vi\'{e}gas
  et~al.}]{10.5555/3454287.3455058}
Coenen A, Reif E, Yuan A, Kim B, Pearce A, Vi\'{e}gas F, et~al. (2019).
\newblock \emph{Visualizing and measuring the geometry of BERT}.
\newblock Curran Associates Inc., Red Hook, NY, USA.

\bibitem[{Colin et~al.(2022)Colin, FEL, Cadene, and Serre}]{colin_what_2022}
Colin J, FEL T, Cadene R, Serre T (2022).
\newblock What {I} {Cannot} {Predict}, {I} {Do} {Not} {Understand}: {A}
  {Human}-{Centered} {Evaluation} {Framework} for {Explainability} {Methods}.
\newblock In: \emph{Advances in {Neural} {Information} {Processing} {Systems}}
  (S~Koyejo, S~Mohamed, A~Agarwal, D~Belgrave, K~Cho, A~Oh, eds.), volume~35,
  2832--2845. Curran Associates, Inc.

\bibitem[{Crabbe et~al.(2021)Crabbe, Qian, Imrie, and van~der
  Schaar}]{crabbe_explaining_2021}
Crabbe J, Qian Z, Imrie F, van~der Schaar M (2021).
\newblock Explaining {Latent} {Representations} with a {Corpus} of {Examples}.
\newblock In: \emph{Advances in {Neural} {Information} {Processing} {Systems}}
  (M~Ranzato, A~Beygelzimer, Y~Dauphin, PS~Liang, JW~Vaughan, eds.), volume~34,
  12154--12166. Curran Associates, Inc.

\bibitem[{Dandl et~al.(2023)Dandl, Casalicchio, Bischl, and
  Bothmann}]{Dandl2023-xl}
Dandl S, Casalicchio G, Bischl B, Bothmann L (2023).
\newblock Interpretable regional descriptors: Hyperbox-based local
  explanations.
\newblock In: \emph{Machine Learning and Knowledge Discovery in Databases:
  Research Track}, Lecture notes in computer science, 479--495. Springer Nature
  Switzerland, Cham.

\bibitem[{Devlin et~al.(2019)Devlin, Chang, Lee, and
  Toutanova}]{devlin-etal-2019-bert}
Devlin J, Chang MW, Lee K, Toutanova K (2019).
\newblock {BERT}: Pre-training of deep bidirectional transformers for language
  understanding.
\newblock In: \emph{Proceedings of the 2019 Conference of the North {A}merican
  Chapter of the Association for Computational Linguistics: Human Language
  Technologies, Volume 1 (Long and Short Papers)} (J~Burstein, C~Doran,
  T~Solorio, eds.), 4171--4186. Association for Computational Linguistics,
  Minneapolis, Minnesota.

\bibitem[{Ding et~al.(2022)Ding, Li, Narasimhan, and Tibshirani}]{Ding2022}
Ding DY, Li S, Narasimhan B, Tibshirani R (2022).
\newblock Cooperative learning for multiview analysis.
\newblock \emph{Proceedings of the National Academy of Sciences}, 119(38).

\bibitem[{Erhan et~al.(2010)Erhan, Bengio, Courville, Manzagol, Vincent, and
  Bengio}]{JMLR:v11:erhan10a}
Erhan D, Bengio Y, Courville A, Manzagol PA, Vincent P, Bengio S (2010).
\newblock Why does unsupervised pre-training help deep learning?
\newblock \emph{Journal of Machine Learning Research}, 11(19): 625--660.

\bibitem[{Fridovich-Keil et~al.(2022)Fridovich-Keil, Yu, Tancik, Chen, Recht,
  and Kanazawa}]{FridovichKeil2022}
Fridovich-Keil S, Yu A, Tancik M, Chen Q, Recht B, Kanazawa A (2022).
\newblock Plenoxels: Radiance fields without neural networks.
\newblock In: \emph{2022 IEEE/CVF Conference on Computer Vision and Pattern
  Recognition (CVPR)}. IEEE.

\bibitem[{Friedman et~al.(2010)Friedman, Hastie, and Tibshirani}]{Friedman2010}
Friedman J, Hastie T, Tibshirani R (2010).
\newblock Regularization paths for generalized linear models via coordinate
  descent.
\newblock \emph{Journal of Statistical Software}, 33(1).

\bibitem[{Friedman(2001)}]{Friedman2001}
Friedman JH (2001).
\newblock Greedy function approximation: A gradient boosting machine.
\newblock \emph{The Annals of Statistics}, 29(5).

\bibitem[{Fumagalli et~al.(2023)Fumagalli, Muschalik, Kolpaczki,
  H{\"u}llermeier, and Hammer}]{Fumagalli2023-vl}
Fumagalli F, Muschalik M, Kolpaczki P, H{\"u}llermeier E, Hammer B (2023).
\newblock {SHAP-IQ}: Unified approximation of any-order shapley interactions.

\bibitem[{Ghorbani et~al.(2019)Ghorbani, Wexler, Zou, and
  Kim}]{10.5555/3454287.3455119}
Ghorbani A, Wexler J, Zou J, Kim B (2019).
\newblock \emph{Towards automatic concept-based explanations}.
\newblock Curran Associates Inc., Red Hook, NY, USA.

\bibitem[{Gosmann et~al.(2017)Gosmann, Anahtar, Handley, Farcasanu, Abu-Ali,
  Bowman et~al.}]{GOSMANN201729}
Gosmann C, Anahtar MN, Handley SA, Farcasanu M, Abu-Ali G, Bowman BA, et~al.
  (2017).
\newblock Lactobacillus-deficient cervicovaginal bacterial communities are
  associated with increased {HIV} acquisition in young {S}outh {A}frican women.
\newblock \emph{Immunity}, 46(1): 29--37.

\bibitem[{Gu et~al.(2019)Gu, Liu, Dolan-Gavitt, and Garg}]{Gu2019}
Gu T, Liu K, Dolan-Gavitt B, Garg S (2019).
\newblock Badnets: Evaluating backdooring attacks on deep neural networks.
\newblock \emph{IEEE Access}, 7: 47230–47244.

\bibitem[{Guidotti(2021)}]{Guidotti2021-bv}
Guidotti R (2021).
\newblock Evaluating local explanation methods on ground truth.
\newblock \emph{Artif. Intell.}, 291(103428): 103428.

\bibitem[{Guidotti(2022)}]{Guidotti2022-zv}
Guidotti R (2022).
\newblock Counterfactual explanations and how to find them: literature review
  and benchmarking.
\newblock \emph{Data Min. Knowl. Discov.}

\bibitem[{Hazimeh et~al.(2020)Hazimeh, Ponomareva, Mol, Tan, and
  Mazumder}]{10.5555/3524938.3525325}
Hazimeh H, Ponomareva N, Mol P, Tan Z, Mazumder R (2020).
\newblock The tree ensemble layer: differentiability meets conditional
  computation.
\newblock In: \emph{Proceedings of the 37th International Conference on Machine
  Learning}, ICML'20. JMLR.org.

\bibitem[{Hedstr{\"o}m et~al.(2023{\natexlab{a}})Hedstr{\"o}m, Bommer,
  Wickstr{\o}m, Samek, Lapuschkin, and H{\"o}hne}]{Hedstrom2023-gd}
Hedstr{\"o}m A, Bommer P, Wickstr{\o}m KK, Samek W, Lapuschkin S, H{\"o}hne MMC
  (2023{\natexlab{a}}).
\newblock The meta-evaluation problem in explainable {AI}: Identifying reliable
  estimators with {MetaQuantus}.

\bibitem[{Hedstr{\"o}m et~al.(2023{\natexlab{b}})Hedstr{\"o}m, Weber,
  Krakowczyk, Bareeva, Motzkus, Samek et~al.}]{JMLR:v24:22-0142}
Hedstr{\"o}m A, Weber L, Krakowczyk D, Bareeva D, Motzkus F, Samek W, et~al.
  (2023{\natexlab{b}}).
\newblock Quantus: An explainable {AI} toolkit for responsible evaluation of
  neural network explanations and beyond.
\newblock \emph{Journal of Machine Learning Research}, 24(34): 1--11.

\bibitem[{Heer(2019)}]{Heer2019}
Heer J (2019).
\newblock Agency plus automation: Designing artificial intelligence into
  interactive systems.
\newblock \emph{Proceedings of the National Academy of Sciences}, 116(6):
  1844–1850.

\bibitem[{Herbinger et~al.(2022)Herbinger, Bischl, and
  Casalicchio}]{Herbinger2022-jb}
Herbinger J, Bischl B, Casalicchio G (2022).
\newblock {REPID}: Regional effect plots with implicit interaction detection.
\newblock 10209--10233.

\bibitem[{Hesse et~al.(2023)Hesse, Schaub-Meyer, and Roth}]{Hesse2023-zq}
Hesse R, Schaub-Meyer S, Roth S (2023).
\newblock {FunnyBirds}: A synthetic vision dataset for a part-based analysis of
  explainable {AI} methods.
\newblock In: \emph{2023 {IEEE/CVF} International Conference on Computer Vision
  ({ICCV})}. IEEE.

\bibitem[{Hewitt and Manning(2019)}]{hewitt-manning-2019-structural}
Hewitt J, Manning CD (2019).
\newblock {A} structural probe for finding syntax in word representations.
\newblock In: \emph{Proceedings of the 2019 Conference of the North {A}merican
  Chapter of the Association for Computational Linguistics: Human Language
  Technologies, Volume 1 (Long and Short Papers)} (J~Burstein, C~Doran,
  T~Solorio, eds.), 4129--4138. Association for Computational Linguistics,
  Minneapolis, Minnesota.

\bibitem[{Holmes(2017)}]{Holmes2017}
Holmes S (2017).
\newblock Statistical proof? the problem of irreproducibility.
\newblock \emph{Bulletin of the American Mathematical Society}, 55(1): 31–55.

\bibitem[{Hooker et~al.(2019)Hooker, Erhan, Kindermans, and
  Kim}]{hooker_benchmark_2019}
Hooker S, Erhan D, Kindermans PJ, Kim B (2019).
\newblock A {Benchmark} for {Interpretability} {Methods} in {Deep} {Neural}
  {Networks}.
\newblock In: \emph{Advances in {Neural} {Information} {Processing} {Systems}}
  (H~Wallach, H~Larochelle, A~Beygelzimer, Fd~Alché-Buc, E~Fox, R~Garnett,
  eds.), volume~32. Curran Associates, Inc.

\bibitem[{Huber(1964)}]{Huber1964}
Huber PJ (1964).
\newblock Robust estimation of a location parameter.
\newblock \emph{The Annals of Mathematical Statistics}, 35(1): 73–101.

\bibitem[{Hutchins et~al.(1985)Hutchins, Hollan, and Norman}]{Hutchins1985}
Hutchins EL, Hollan JD, Norman DA (1985).
\newblock Direct manipulation interfaces.
\newblock \emph{Human–Computer Interaction}, 1(4): 311–338.

\bibitem[{Jeganathan et~al.(2018)Jeganathan, Callahan, Proctor, Relman, and
  Holmes}]{block_bootstrap}
Jeganathan P, Callahan BJ, Proctor DM, Relman DA, Holmes SP (2018).
\newblock The block bootstrap method for longitudinal microbiome data.

\bibitem[{Jeyakumar et~al.(2020)Jeyakumar, Noor, Cheng, Garcia, and
  Srivastava}]{jeyakumar_how_2020}
Jeyakumar JV, Noor J, Cheng YH, Garcia L, Srivastava M (2020).
\newblock How {Can} {I} {Explain} {This} to {You}? {An} {Empirical} {Study} of
  {Deep} {Neural} {Network} {Explanation} {Methods}.
\newblock In: \emph{Advances in {Neural} {Information} {Processing} {Systems}},
  volume~33, 4211--4222. Curran Associates, Inc.

\bibitem[{Karimi et~al.(2022)Karimi, Barthe, Sch\"{o}lkopf, and
  Valera}]{Karimi2022}
Karimi AH, Barthe G, Sch\"{o}lkopf B, Valera I (2022).
\newblock A survey of algorithmic recourse: Contrastive explanations and
  consequential recommendations.
\newblock \emph{ACM Computing Surveys}, 55(5): 1–29.

\bibitem[{Kim(2022)}]{medium_beyond_nodate}
Kim B (2022).
\newblock Beyond {Interpretability}: {Developing} a {Language} to {Shape} {Our}
  {Relationships} with {AI}.

\bibitem[{Kodikara et~al.(2022)Kodikara, Ellul, and L\^e~Cao}]{Kodikara2022}
Kodikara S, Ellul S, L\^e~Cao KA (2022).
\newblock Statistical challenges in longitudinal microbiome data analysis.
\newblock \emph{Briefings in Bioinformatics}, 23(4).

\bibitem[{Koh and Liang(2017)}]{10.5555/3305381.3305576}
Koh PW, Liang P (2017).
\newblock Understanding black-box predictions via influence functions.
\newblock In: \emph{Proceedings of the 34th International Conference on Machine
  Learning - Volume 70}, ICML'17, 1885–1894. JMLR.org.

\bibitem[{Koh et~al.(2020)Koh, Nguyen, Tang, Mussmann, Pierson, Kim
  et~al.}]{50351}
Koh PW, Nguyen T, Tang YS, Mussmann S, Pierson E, Kim B, et~al. (2020).
\newblock Concept bottleneck models.

\bibitem[{Kolesnikov et~al.(2021)Kolesnikov, Dosovitskiy, Weissenborn, Heigold,
  Uszkoreit, Beyer et~al.}]{50650}
Kolesnikov A, Dosovitskiy A, Weissenborn D, Heigold G, Uszkoreit J, Beyer L,
  et~al. (2021).
\newblock An image is worth 16x16 words: Transformers for image recognition at
  scale.

\bibitem[{Kostic et~al.(2015)Kostic, Gevers, Siljander, Vatanen,
  Hy\"{o}tyl\"{a}inen, H\"{a}m\"{a}l\"{a}inen et~al.}]{Kostic2015}
Kostic AD, Gevers D, Siljander H, Vatanen T, Hy\"{o}tyl\"{a}inen T,
  H\"{a}m\"{a}l\"{a}inen AM, et~al. (2015).
\newblock The dynamics of the human infant gut microbiome in development and in
  progression toward type 1 diabetes.
\newblock \emph{Cell Host and Microbe}, 17(2): 260–273.

\bibitem[{Krishnan(2019)}]{Krishnan2019}
Krishnan M (2019).
\newblock Against interpretability: a critical examination of the
  interpretability problem in machine learning.
\newblock \emph{Philosophy and Technology}, 33(3): 487–502.

\bibitem[{Kundaliya(2023)}]{kundaliya2023}
Kundaliya D (2023).
\newblock Computing - incisive media: Google ai chatbot bard gives wrong answer
  in its first demo.
\newblock \emph{Computing}.
\newblock Nom - OpenAI; Copyright - Copyright Newstex Feb 9, 2023; Dernière
  mise à jour - 2023-11-30.

\bibitem[{Lee(2021)}]{lee_transformers_nodate}
Lee JS (2021).
\newblock Transformers: a {Primer}.

\bibitem[{Lipton(2018)}]{Lipton2018}
Lipton ZC (2018).
\newblock The mythos of model interpretability: In machine learning, the
  concept of interpretability is both important and slippery.
\newblock \emph{Queue}, 16(3): 31–57.

\bibitem[{Liu et~al.(2021)Liu, Khandagale, White, and Neiswanger}]{Liu2021-ao}
Liu Y, Khandagale S, White C, Neiswanger W (2021).
\newblock Synthetic benchmarks for scientific research in explainable machine
  learning.

\bibitem[{Loh(2014)}]{Loh2014}
Loh W (2014).
\newblock Fifty years of classification and regression trees.
\newblock \emph{International Statistical Review}, 82(3): 329–348.

\bibitem[{Lundberg et~al.(2020)Lundberg, Erion, Chen, DeGrave, Prutkin, Nair
  et~al.}]{Lundberg2020-cu}
Lundberg SM, Erion G, Chen H, DeGrave A, Prutkin JM, Nair B, et~al. (2020).
\newblock From local explanations to global understanding with explainable {AI}
  for trees.
\newblock \emph{Nat. Mach. Intell.}, 2(1): 56--67.

\bibitem[{Lundberg and Lee(2017)}]{10.5555/3295222.3295230}
Lundberg SM, Lee SI (2017).
\newblock A unified approach to interpreting model predictions.
\newblock In: \emph{Proceedings of the 31st International Conference on Neural
  Information Processing Systems}, NIPS'17, 4768–4777. Curran Associates
  Inc., Red Hook, NY, USA.

\bibitem[{Lundstrom et~al.(2022)Lundstrom, Huang, and
  Razaviyayn}]{pmlr-v162-lundstrom22a}
Lundstrom DD, Huang T, Razaviyayn M (2022).
\newblock A rigorous study of integrated gradients method and extensions to
  internal neuron attributions.
\newblock In: \emph{Proceedings of the 39th International Conference on Machine
  Learning} (K~Chaudhuri, S~Jegelka, L~Song, C~Szepesvari, G~Niu, S~Sabato,
  eds.), volume 162 of \emph{Proceedings of Machine Learning Research},
  14485--14508. PMLR.

\bibitem[{Ma et~al.(2024)Ma, Lai, Zhang, Chen, Hamilton, Ljubenkov
  et~al.}]{Ma2024-nt}
Ma J, Lai V, Zhang Y, Chen C, Hamilton P, Ljubenkov D, et~al. (2024).
\newblock {OpenHEXAI}: An open-source framework for human-centered evaluation
  of explainable machine learning.

\bibitem[{Mehrabi et~al.(2021)Mehrabi, Morstatter, Saxena, Lerman, and
  Galstyan}]{Mehrabi2021}
Mehrabi N, Morstatter F, Saxena N, Lerman K, Galstyan A (2021).
\newblock A survey on bias and fairness in machine learning.
\newblock \emph{ACM Computing Surveys}, 54(6): 1–35.

\bibitem[{Mothilal et~al.(2020)Mothilal, Sharma, and Tan}]{Mothilal2020-zz}
Mothilal RK, Sharma A, Tan C (2020).
\newblock Explaining machine learning classifiers through diverse
  counterfactual explanations.
\newblock In: \emph{Proceedings of the 2020 Conference on Fairness,
  Accountability, and Transparency}. ACM, New York, NY, USA.

\bibitem[{Murdoch et~al.(2019)Murdoch, Singh, Kumbier, Abbasi-Asl, and
  Yu}]{Murdoch2019}
Murdoch WJ, Singh C, Kumbier K, Abbasi-Asl R, Yu B (2019).
\newblock Definitions, methods, and applications in interpretable machine
  learning.
\newblock \emph{Proceedings of the National Academy of Sciences}, 116(44):
  22071–22080.

\bibitem[{Nannini et~al.(2023)Nannini, Balayn, and Smith}]{Nannini2023-ep}
Nannini L, Balayn A, Smith AL (2023).
\newblock Explainability in {AI} policies: A critical review of communications,
  reports, regulations, and standards in the {EU}, {US}, and {UK}.
\newblock In: \emph{2023 {ACM} Conference on Fairness, Accountability, and
  Transparency}. ACM, New York, NY, USA.

\bibitem[{Ngiam et~al.(2011)Ngiam, Khosla, Kim, Nam, Lee, and
  Ng}]{10.5555/3104482.3104569}
Ngiam J, Khosla A, Kim M, Nam J, Lee H, Ng AY (2011).
\newblock Multimodal deep learning.
\newblock In: \emph{Proceedings of the 28th International Conference on
  International Conference on Machine Learning}, ICML'11, 689–696. Omnipress,
  Madison, WI, USA.

\bibitem[{Nguyen and Holmes(2019)}]{Nguyen2019}
Nguyen LH, Holmes S (2019).
\newblock Ten quick tips for effective dimensionality reduction.
\newblock \emph{PLOS Computational Biology}, 15(6): e1006907.

\bibitem[{Nussberger et~al.(2022)Nussberger, Luo, Celis, and
  Crockett}]{Nussberger2022}
Nussberger AM, Luo L, Celis LE, Crockett MJ (2022).
\newblock Public attitudes value interpretability but prioritize accuracy in
  artificial intelligence.
\newblock \emph{Nature Communications}, 13(1).

\bibitem[{Oppermann and Munzner(2022)}]{9555620}
Oppermann M, Munzner T (2022).
\newblock Vizsnippets: Compressing visualization bundles into representative
  previews for browsing visualization collections.
\newblock \emph{IEEE Transactions on Visualization and Computer Graphics},
  28(1): 747--757.

\bibitem[{Pennington et~al.(2014)Pennington, Socher, and
  Manning}]{Pennington2014}
Pennington J, Socher R, Manning C (2014).
\newblock Glove: Global vectors for word representation.
\newblock In: \emph{Proceedings of the 2014 Conference on Empirical Methods in
  Natural Language Processing (EMNLP)}. Association for Computational
  Linguistics.

\bibitem[{Poursabzi-Sangdeh et~al.(2021)Poursabzi-Sangdeh, Goldstein, Hofman,
  Wortman~Vaughan, and Wallach}]{Poursabzi-Sangdeh2021-ws}
Poursabzi-Sangdeh F, Goldstein DG, Hofman JM, Wortman~Vaughan JW, Wallach H
  (2021).
\newblock Manipulating and measuring model interpretability.
\newblock In: \emph{Proceedings of the 2021 {CHI} Conference on Human Factors
  in Computing Systems}. ACM, New York, NY, USA.

\bibitem[{Raghu et~al.(2021)Raghu, Unterthiner, Kornblith, Zhang, and
  Dosovitskiy}]{raghu_vision_2021}
Raghu M, Unterthiner T, Kornblith S, Zhang C, Dosovitskiy A (2021).
\newblock Do {Vision} {Transformers} {See} {Like} {Convolutional} {Neural}
  {Networks}?
\newblock In: \emph{Advances in {Neural} {Information} {Processing} {Systems}}
  (M~Ranzato, A~Beygelzimer, Y~Dauphin, PS~Liang, JW~Vaughan, eds.), volume~34,
  12116--12128. Curran Associates, Inc.

\bibitem[{Raghu et~al.(2019)Raghu, Zhang, Kleinberg, and
  Bengio}]{10.5555/3454287.3454588}
Raghu M, Zhang C, Kleinberg J, Bengio S (2019).
\newblock \emph{Transfusion: understanding transfer learning for medical
  imaging}.
\newblock Curran Associates Inc., Red Hook, NY, USA.

\bibitem[{Resnick and Varian(1997)}]{resnick1997recommender}
Resnick P, Varian HR (1997).
\newblock Recommender systems.
\newblock \emph{Communications of the ACM}, 40(3): 56--58.

\bibitem[{Ribeiro et~al.(2016)Ribeiro, Singh, and
  Guestrin}]{10.1145/2939672.2939778}
Ribeiro MT, Singh S, Guestrin C (2016).
\newblock ``{W}hy should {I} trust you?'': {E}xplaining the predictions of any
  classifier.
\newblock In: \emph{Proceedings of the 22nd ACM SIGKDD International Conference
  on Knowledge Discovery and Data Mining}, KDD '16, 1135–1144. Association
  for Computing Machinery, New York, NY, USA.

\bibitem[{Rong et~al.(2024)Rong, Leemann, Nguyen, Fiedler, Qian, Unhelkar
  et~al.}]{Rong2024-ss}
Rong Y, Leemann T, Nguyen TT, Fiedler L, Qian P, Unhelkar V, et~al. (2024).
\newblock Towards human-centered explainable {AI}: A survey of user studies for
  model explanations.
\newblock \emph{IEEE Trans. Pattern Anal. Mach. Intell.}, 46(4): 2104--2122.

\bibitem[{Ross et~al.(2017)Ross, Hughes, and Doshi-Velez}]{Ross2017-ey}
Ross AS, Hughes MC, Doshi-Velez F (2017).
\newblock Right for the right reasons: Training differentiable models by
  constraining their explanations.
\newblock In: \emph{Proceedings of the {Twenty-Sixth} International Joint
  Conference on Artificial Intelligence}. International Joint Conferences on
  Artificial Intelligence Organization, California.

\bibitem[{Rudin(2019)}]{Rudin2019}
Rudin C (2019).
\newblock Stop explaining black box machine learning models for high stakes
  decisions and use interpretable models instead.
\newblock \emph{Nature Machine Intelligence}, 1(5): 206–215.

\bibitem[{Sankaran and Holmes(2023)}]{Sankaran2023}
Sankaran K, Holmes SP (2023).
\newblock Generative models: An interdisciplinary perspective.
\newblock \emph{Annual Review of Statistics and Its Application}, 10(1):
  325–352.

\bibitem[{Sedlmair et~al.(2012)Sedlmair, Meyer, and Munzner}]{6327248}
Sedlmair M, Meyer M, Munzner T (2012).
\newblock Design study methodology: Reflections from the trenches and the
  stacks.
\newblock \emph{IEEE Transactions on Visualization and Computer Graphics},
  18(12): 2431--2440.

\bibitem[{Selvaraju et~al.(2017)Selvaraju, Cogswell, Das, Vedantam, Parikh, and
  Batra}]{8237336}
Selvaraju RR, Cogswell M, Das A, Vedantam R, Parikh D, Batra D (2017).
\newblock Grad-cam: Visual explanations from deep networks via gradient-based
  localization.
\newblock In: \emph{2017 IEEE International Conference on Computer Vision
  (ICCV)}, 618--626.

\bibitem[{Silverman et~al.(2018)Silverman, Shenhav, Halperin, Mukherjee, and
  David}]{Silverman2018}
Silverman JD, Shenhav L, Halperin E, Mukherjee S, David LA (2018).
\newblock Statistical considerations in the design and analysis of longitudinal
  microbiome studies.

\bibitem[{Simonyan et~al.(2014)Simonyan, Vedaldi, and Zisserman}]{Simonyan14a}
Simonyan K, Vedaldi A, Zisserman A (2014).
\newblock Deep inside convolutional networks: Visualising image classification
  models and saliency maps.
\newblock In: \emph{Workshop at International Conference on Learning
  Representations}.

\bibitem[{Sokol and Flach(2020)}]{Sokol2020-de}
Sokol K, Flach P (2020).
\newblock One explanation does not fit all.
\newblock \emph{KI - K{\"u}nstl. Intell.}, 34(2): 235--250.

\bibitem[{Sturmfels et~al.(2020)Sturmfels, Lundberg, and Lee}]{Sturmfels2020}
Sturmfels P, Lundberg S, Lee SI (2020).
\newblock Visualizing the impact of feature attribution baselines.
\newblock \emph{Distill}, 5(1).

\bibitem[{Sundararajan et~al.(2017)Sundararajan, Taly, and
  Yan}]{10.5555/3305890.3306024}
Sundararajan M, Taly A, Yan Q (2017).
\newblock Axiomatic attribution for deep networks.
\newblock In: \emph{Proceedings of the 34th International Conference on Machine
  Learning - Volume 70}, ICML'17, 3319–3328. JMLR.org.

\bibitem[{Teh(2019)}]{teh_statistical_nodate}
Teh YW (2019).
\newblock On {Statistical} {Thinking} in {Deep} {Learning} {A} {Blog} {Post}.
\newblock \emph{IMS Medallion Lecture}.

\bibitem[{Tufte(2001)}]{Tufte2001-xs}
Tufte ER (2001).
\newblock \emph{The visual display of quantitative information}.
\newblock Graphics Press, Cheshire, CT, 2 edition.

\bibitem[{Tukey(1959)}]{Tukey:1959:SSC}
Tukey JW (1959).
\newblock A survey of sampling from contaminated distributions.
\newblock \emph{STRG Technical report}, (33): 57.

\bibitem[{{Van Boxtel, G.J.M., et al.}(2021)}]{gsignal}
{Van Boxtel, GJM, et al} (2021).
\newblock \emph{{gsignal}: Signal processing}.

\bibitem[{Vaswani et~al.(2017)Vaswani, Shazeer, Parmar, Uszkoreit, Jones, Gomez
  et~al.}]{10.5555/3295222.3295349}
Vaswani A, Shazeer N, Parmar N, Uszkoreit J, Jones L, Gomez AN, et~al. (2017).
\newblock Attention is all you need.
\newblock In: \emph{Proceedings of the 31st International Conference on Neural
  Information Processing Systems}, NIPS'17, 6000–6010. Curran Associates
  Inc., Red Hook, NY, USA.

\bibitem[{Wang et~al.(2022)Wang, Shen, Peng, Zhang, Xiao, Liu
  et~al.}]{Wang2022-aj}
Wang L, Shen Y, Peng S, Zhang S, Xiao X, Liu H, et~al. (2022).
\newblock A fine-grained interpretability evaluation benchmark for neural
  {NLP}.
\newblock In: \emph{Proceedings of the 26th Conference on Computational Natural
  Language Learning ({CoNLL})}. Association for Computational Linguistics,
  Stroudsburg, PA, USA.

\bibitem[{Williamson and Feng(2020)}]{pmlr-v119-williamson20a}
Williamson B, Feng J (2020).
\newblock Efficient nonparametric statistical inference on population feature
  importance using shapley values.
\newblock In: \emph{Proceedings of the 37th International Conference on Machine
  Learning} (HD~III, A~Singh, eds.), volume 119 of \emph{Proceedings of Machine
  Learning Research}, 10282--10291. PMLR.

\bibitem[{Wongsuphasawat et~al.(2018)Wongsuphasawat, Smilkov, Wexler, Wilson,
  Man{\'e}, Fritz et~al.}]{Wongsuphasawat2018VisualizingDG}
Wongsuphasawat K, Smilkov D, Wexler J, Wilson J, Man{\'e} D, Fritz D, et~al.
  (2018).
\newblock Visualizing dataflow graphs of deep learning models in tensorflow.
\newblock \emph{IEEE Transactions on Visualization and Computer Graphics}, 24:
  1--12.

\bibitem[{Wu et~al.(2021)Wu, Ribeiro, Heer, and Weld}]{Wu2021}
Wu T, Ribeiro MT, Heer J, Weld D (2021).
\newblock Polyjuice: Generating counterfactuals for explaining, evaluating, and
  improving models.
\newblock In: \emph{Proceedings of the 59th Annual Meeting of the Association
  for Computational Linguistics and the 11th International Joint Conference on
  Natural Language Processing (Volume 1: Long Papers)}. Association for
  Computational Linguistics.

\bibitem[{Xie et~al.(2022)Xie, Raghunathan, Liang, and
  Ma}]{DBLP:conf/iclr/XieRL022}
Xie SM, Raghunathan A, Liang P, Ma T (2022).
\newblock An explanation of in-context learning as implicit bayesian inference.
\newblock In: \emph{The Tenth International Conference on Learning
  Representations, {ICLR} 2022, Virtual Event, April 25-29, 2022}.
  OpenReview.net.

\bibitem[{Xin et~al.(2022)Xin, Zhong, Chen, Takagi, Seltzer, and
  Rudin}]{Xin2022-ie}
Xin R, Zhong C, Chen Z, Takagi T, Seltzer M, Rudin C (2022).
\newblock Exploring the whole rashomon set of sparse decision trees.
\newblock \emph{ArXiv}, arXiv:2209.08040.

\bibitem[{Yosinski et~al.(2015)Yosinski, Clune, Nguyen, Fuchs, and
  Lipson}]{Yosinski2015UnderstandingNN}
Yosinski J, Clune J, Nguyen AM, Fuchs TJ, Lipson H (2015).
\newblock Understanding neural networks through deep visualization.
\newblock \emph{ArXiv}, abs/1506.06579.

\bibitem[{Yuksekgonul et~al.(2023)Yuksekgonul, Wang, and
  Zou}]{yuksekgonul2023posthoc}
Yuksekgonul M, Wang M, Zou J (2023).
\newblock Post-hoc concept bottleneck models.
\newblock In: \emph{The Eleventh International Conference on Learning
  Representations}.

\bibitem[{Zeiler and Fergus(2013)}]{Zeiler2013VisualizingAU}
Zeiler MD, Fergus R (2013).
\newblock Visualizing and understanding convolutional networks.
\newblock \emph{ArXiv}, abs/1311.2901.

\bibitem[{Zeng et~al.(2016)Zeng, Ustun, and Rudin}]{Zeng2016}
Zeng J, Ustun B, Rudin C (2016).
\newblock Interpretable classification models for recidivism prediction.
\newblock \emph{Journal of the Royal Statistical Society Series A: Statistics
  in Society}, 180(3): 689–722.

\bibitem[{Zhong et~al.(2023)Zhong, Chen, Liu, Seltzer, and
  Rudin}]{Zhong2023-zi}
Zhong C, Chen Z, Liu J, Seltzer M, Rudin C (2023).
\newblock Exploring and interacting with the set of good sparse generalized
  additive models.

\bibitem[{Zhou et~al.(2022)Zhou, Booth, Ribeiro, and Shah}]{Zhou2022-lb}
Zhou Y, Booth S, Ribeiro MT, Shah J (2022).
\newblock Do feature attribution methods correctly attribute features?
\newblock \emph{Proc. Conf. AAAI Artif. Intell.}, 36(9): 9623--9633.

\bibitem[{Zou et~al.(2023)Zou, Yang, Zhang, Li, Li, Wang
  et~al.}]{zou_segment_2023}
Zou X, Yang J, Zhang H, Li F, Li L, Wang J, et~al. (2023).
\newblock Segment {Everything} {Everywhere} {All} at {Once}.
\newblock In: \emph{Advances in {Neural} {Information} {Processing} {Systems}}
  (A~Oh, T~Naumann, A~Globerson, K~Saenko, M~Hardt, S~Levine, eds.), volume~36,
  19769--19782. Curran Associates, Inc.

\end{thebibliography}

\end{document}


\begin{table}[]
  \centering
  \begin{tabular}{|p{1.4cm}p{3.5cm}R{1cm}R{2cm}R{3cm}R{1.8cm}|}
  \hline
  \textbf{Data} & \textbf{Model} & \textbf{$N$}  & \textbf{In-Sample Accuracy} & \textbf{Out-of-Sample Accuracy} & \textbf{Training Time (s)}\\ \hline
  Original & Sparse Logistic Reg. & 500  & 81.2\%         &   78.0\% &  3.9    \\
  &  & 1000  & 82.1\%         &   81.0\% &  15.9   \\
  &  & 5000  & 78.1\%         &   77.9\% &  22.9    \\
  &  & 10000  & \%         &   \% &      \\
  & Decision Tree & 500 & 91.6\%        & 70.2\%  & 95.9         \\
  & & 1000 & 91.3\% & 76.5\%  & 497.9         \\
  & & 5000 & 91.4\% & 75.7\%  &    3255.3   \\
  & & 10000 & \% & \%  &      \\
  & Transformer    &  500 & 100\% & 86.4\%  & 91.2  \\ 
  & &  1000 & 90.7\%  &  76.0\%  & 246.6 \\ 
  & &  5000 & 96.6\% & 86.4\%  & 1167.6 \\ 
  & &  10000 & 87.6\% &   87.7\%  & 1964.4 \\ 
  & Concept Bottleneck & 500 & 98.4\%  &  84.8\% &  110.9   \\
  & & 1000 & 99.2\%   & 73.2\% &  211.6  \\
  & & 5000 & 90.4\%  &   88.2\% & 892.8 \\
  & & 10000 & 92.5\%   &  89.0\% &  1854.0   \\ \hline
  Featurized & Sparse Logistic Reg. & 500 & 92.6\%  & 87.8\% &  0.4  \\
  &  & 1000 & 88.8\% &   85.4\% &  8.1  \\
  &  & 5000 & 86.5\% &   85.3\% &  32.7  \\
  &  & 10000 & \% &   \% &  0.4  \\
  & Decision Tree & 500 & 74.4\%     &  69.6\%  &  1.8 \\
  & & 1000 & 87.9\% &  70.6\%  &  14.7 \\
  & & 5000 & 90.2\%  &  75.1\%  &  185.5 \\
  & & 10000 & \%     &  \%  &   \\
  \hline
  \end{tabular}
  \caption{In and out-of-sample accuracies of models applied to the simulation
  study. For cross-validated models, training time was averaged across folds.
  Training time was computed on a 2022 MacBook Pro with 16GB RAM and an M2 GPU.
  Models differ in their inherent interpretability, manual feature curation
  effort, and generalization performance. Deep learning models can overfit the
  training sample while still generalizing well. Intrinsically interpretable models
  can be substantially improved through effective featurization.}
  \label{tab:tab1}
\end{table}